\pdfoutput=1
\documentclass{article}

\usepackage{arxiv}
\usepackage{amsmath,amssymb,amsfonts,amsthm,mathtools}
\usepackage[utf8]{inputenc}
\usepackage[T1]{fontenc}
\usepackage[numbers,sort&compress]{natbib}
\usepackage{hyperref}
\usepackage{url}

\usepackage{booktabs}
\usepackage{array}
\usepackage{multirow}
\usepackage{microtype}
\usepackage{xcolor}
\usepackage{tikz}
\usetikzlibrary{arrows.meta,calc}
\usepackage{graphicx}
\usepackage{float}
\usepackage{placeins}
\usepackage{needspace}
\usepackage{caption}
\usepackage{enumitem}
\usepackage{listings}
\usepackage[capitalize,noabbrev]{cleveref}

\setlist{nosep,leftmargin=1.4em}
\hypersetup{colorlinks=true,linkcolor=blue!60!black,citecolor=blue!60!black,urlcolor=blue!60!black,pdftitle={Certified Against Which Oracle? Execution Labels Set the Reported Risk of Conformal Abstention for Text-to-SQL},pdfauthor={Jiamiao Liu, Dewen Qiao, Yu Zhang, Xuetao Chen}}

\newcommand{\cellA}{\mathrm{A}}
\newcommand{\cellB}{\mathrm{B}}
\newcommand{\cellC}{\mathrm{C}}
\newcommand{\cellD}{\mathrm{D}}
\newcommand{\GAP}{\mathrm{GAP}}
\newcommand{\DmA}{\mathrm{D}{-}\mathrm{A}}

\newcommand{\weak}{\mathrm{w}}
\newcommand{\suite}{\mathrm{s}}
\newcommand{\score}{s}
\newcommand{\thr}{\lambda}
\newcommand{\risk}{R}

\newcommand{\dataurl}{\url{https://github.com/yahiko-l/text-to-sql-oracle-certificates}}

\renewcommand{\shorttitle}{Certified Against Which Oracle? The Risk of Conformal Abstention for Text-to-SQL}
\renewcommand{\undertitle}{Preprint}

\title{Certified Against Which Oracle?\\Execution Labels Set the Reported Risk\\of Conformal Abstention for Text-to-SQL}

\author{%
  Jiamiao Liu \quad Dewen Qiao \quad Yu Zhang \quad Xuetao Chen\thanks{Corresponding author. Email: \texttt{xuetao@tmmu.edu.cn}} \\
  Department of Information, Xinqiao Hospital \\
  Army Medical University (Third Military Medical University) \\
  Chongqing 400037, China
}

\date{}

\begin{document}

\maketitle

\begin{abstract}
A conformal abstention certificate for text-to-SQL is only as truthful as the correctness labels it is calibrated on. The uncertainty pipelines that read confidence off execution consistency take those labels from the single database a benchmark ships, an oracle known to be lenient. We run a preregistered intervention on Spider-Realistic, swapping that database for the benchmark's distilled multi-instance test suite. Across four SQL-specialist checkpoints and two split schemes, the swap raises the certificate's held-out risk 2.73 to 10.23 points above the risk its own labels report. Neither oracle reports the risk experts assign. Under blinded labels from two SQL experts, a certificate calibrated at a nominal 0.10 carries 20.0 and 17.2 points of risk on two checkpoints. The stricter oracle errs in both directions: most of the answers it rejects are not judged wrong, and some of those it accepts are. An AI-assigned census of what it rejects finds a semantic error in a quarter to a third of them, depending on the population. It attributes most of the rest to underspecified questions, synthetic instances or suspected reference-query defects, a flag supported by a preregistered blinded expert audit. The oracle also decides how a confidence score is judged. Every execution-consistency score looks better under the labels of the oracle that built its clusters, in 16 of 16 combinations. Under expert labels, building such a score on suite clusters instead of shipped-database clusters raises its area under the ROC curve (AUROC) by 6.96 points on one checkpoint and 1.53 on the other. On the second, the expert interval excludes the 8.3 points the suite labels report. A certificate should be reported with both oracles, and an oracle-relative difference read as semantic risk only after the benchmark is audited. A consistency score should be evaluated under an oracle that did not build it.

\end{abstract}

\keywords{text-to-SQL \and uncertainty quantification \and conformal risk control \and selective prediction \and benchmark validity \and execution-based evaluation}

\section{Introduction}
\label{sec:intro}

A text-to-SQL system answers a natural-language question by generating a query and returning what the database retrieves, and nothing in the result of a wrong query marks it as wrong \citep{wu2026never}. Abstaining on the questions the system is unsure of is one remedy, and a conformal risk certificate makes that abstention a specific promise. Conformal risk control selects a threshold on a calibration set so that the expected value of a bounded, monotone loss is at most a nominal level $\alpha$ on exchangeable test data \citep{angelopoulos2024crc}. For abstention the loss is one if the system answers and is wrong, so the promise is that, over questions drawn in the same way as the calibration questions, the marginal probability of returning a wrong answer, with abstentions counted as zero loss, is at most $\alpha$. That promise is only as good as the definition of ``wrong'' used to calibrate it. In the text-to-SQL uncertainty pipelines of \cref{sec:related} that read confidence off execution consistency, correctness is judged on one database per schema, by comparing the result of a generated query with the expected one, and where candidates are clustered by whether they return the same result, that database also builds the clusters the confidence score is read off \citep{chen2026endtoend,maleki2025confidence,richardson2026predicts}.

\begin{figure}[t]
    \centering
    \includegraphics[width=\textwidth]{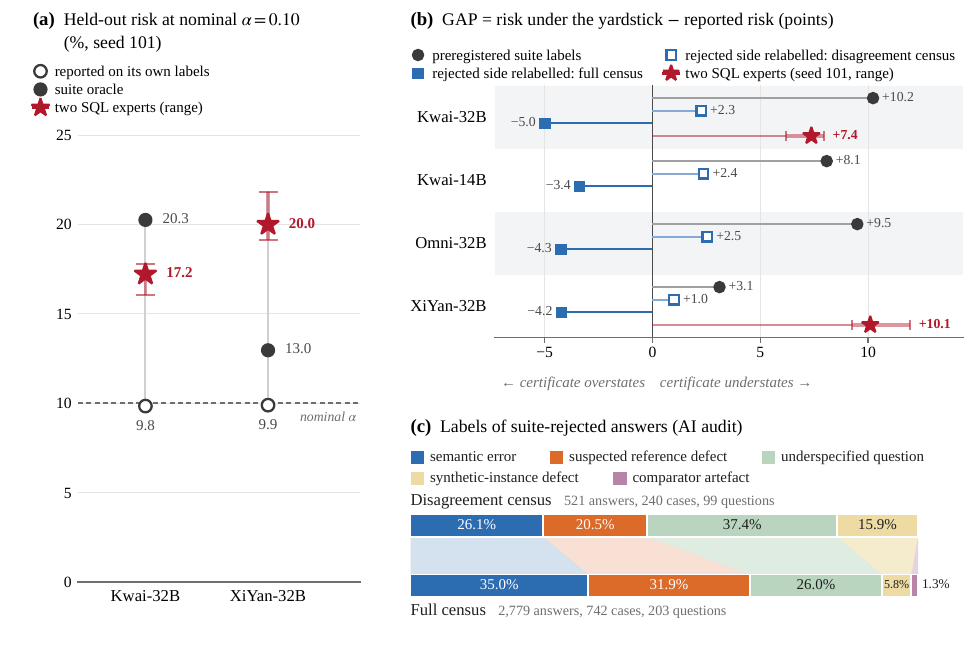}
    \caption{The oracle that labels the calibration set decides the risk the certificate reports, and neither oracle the benchmark publishes reports the risk two SQL experts assign. (a) Held-out risk of the current-practice certificate at nominal risk 0.10 under question splits, at generation seed 101, on the two checkpoints whose distinct answers at that seed were judged by two SQL experts blind to both oracles (\cref{sec:expertrisk}): under its own labels, under the suite oracle, and under the expert labels, where the star falls back to the suite label on the questions the experts left unjudged and the bar spans the nine conventions of \cref{tab:human_gap}. The shipped database understates the expert-judged risk on both checkpoints, and the suite understates it on one and overstates it on the other. (b) $\GAP$, the held-out risk the current-practice certificate carries under a yardstick minus the risk it reports under its own labels, at nominal risk 0.10 under question splits: three-seed means for four checkpoints under the preregistered suite labels and after relabelling only the rejected side with each census of (c) (narrow convention, \cref{sec:reshapes}), and the expert labels of (a) at seed 101 on the two checkpoints they judged, which cover both sides of the oracle and whose bar spans the nine conventions. Relabelling the rejected side with the full census reverses the sign for every checkpoint; the expert labels fix it as positive on both checkpoints they cover. (c) Labels of every answer the multi-instance suite oracle rejects, assigned by a two-pass adjudicated AI audit; a preregistered blinded audit by two SQL experts supports the suspected reference-defect label at the question level (\cref{app:humanaudit}), and the other labels are AI-assigned. The disagreement census holds the answers the shipped database accepts and the suite rejects, the full census every answer of the current-practice or fully multi-instance cell that the suite rejects; the former is nested inside the latter, so the two bars are not independent replications.}
    \label{fig:hero}
\end{figure}

That oracle has been known to be lenient for years. Text-to-SQL benchmarks moved from exact match to execution accuracy because many correct queries do not match the reference string, and execution on the benchmark's database counts such queries as correct \citep{yu2018spider,li2023bird}; execution on one database is lenient in the other direction, because two queries with different meanings can return the same table on the data that happens to be there. \citet{zhong2020testsuite} addressed this with distilled test suites, a small set of databases per schema, distilled from many random ones, on which two queries must agree before they count as equivalent, and the Spider leaderboard adopted test-suite accuracy as its official metric later that year; later work has shown by formal verification and by synthesised databases that single-database execution accepts semantically different queries \citep{klopfenstein2025spotit,habibollah2026synsql}. Two pilot measurements on official Spider dev (\cref{sec:oracles}) set the scale: the shipped databases make the reference queries of 688 pairs of different questions return identical results, of which 0.44 percent survive the suite, and they accept about 7 percent of the reference-query mutations that change the result on some suite instance. Yet the uncertainty pipelines that read confidence off execution consistency still build their signals and judge correctness on one database per schema, and none of them measures what the oracle costs at the certificate layer, or checks whether the stricter oracle's rejections are semantic errors at all. The second question is one of construct validity, whether a correctness label produced by executing a reference query measures whether the question was answered, and it decides more of the reported risk than the first. A third question follows from the same arrangement: where candidates are clustered by execution result, the oracle that supplies the labels also builds the clusters the confidence score is read from, so an evaluation of that score under that oracle cannot separate a score that predicts correctness better from one that agrees with the oracle that defined it, and none of the pipelines that cluster candidates evaluates such a score under an oracle that did not build its clusters.

\paragraph{What we do.} We run a preregistered two-by-two intervention on the oracle of the simplest execution-based certificate: sample fifty candidates per question, partition them into execution-equivalence classes, score a question by the mass of its largest class, and select an abstention threshold by split-half conformal risk control at $\alpha = 0.10$. The partition and the correctness labels can each use the shipped database or every instance of the distilled suite, giving four cells; cell $\cellA$ follows current practice in using the shipped database for both, and cell $\cellD$ is fully multi-instance. $\GAP$ is the held-out risk that cell $\cellA$ carries under the suite oracle minus the held-out risk it carries under its own weak labels, at the threshold its calibration selected; it is the understatement a practitioner who evaluates on the shipped database never sees. $\DmA$ is the suite-oracle risk of cell $\cellD$ minus that of cell $\cellA$. The panel is four SQL-specialist checkpoints from two Qwen lineages, evaluated on all 508 questions of Spider-Realistic with three generation seeds each, 200 split-halves under two split schemes and a false-schema negative control. Post hoc, the intervention is repeated for five published confidence scores under the identical certificate, every answer the suite rejects is labelled in a two-pass adjudicated audit by one AI reviewer family, and the certificate is re-scored under nested relabelling conventions. Two preregistered blinded audits by two SQL experts then supply labels no oracle produced: one tests the census's suspected reference-defect label at the question level, the other judges every distinct answer two checkpoints return at one seed, blind to both oracles.

\paragraph{What we find.} Which oracle labels the calibration set decides the risk the certificate reports, and neither oracle reports the risk two SQL experts assign (\cref{fig:hero}a). In the preregistered analysis the swap raises the held-out risk of the current-practice certificate 2.73 to 10.23 points above the risk it reports under its own labels, on every checkpoint, seed and split scheme, and the fully multi-instance cell carries less suite-oracle risk in every case, by more than the preregistered three points on Kwai-32B, Kwai-14B and Omni-32B but not on XiYan-32B; the post-hoc extension to six confidence scores gives the same picture for every score (\cref{fig:hero}b, preregistered suite labels).

Most of what the suite rejects is not a semantic error. The AI audit assigns that label to 26.1 percent of the answers the shipped database accepts and the suite rejects, and to 35.0 percent of all rejected answers, and attributes the rest to underspecified questions, synthetic instances, comparator artefacts or suspected defects in the benchmark's own reference queries (\cref{fig:hero}c). A preregistered blinded expert audit supports the reference-defect flag: two SQL experts judge 54.8 percent of the flagged questions defective against 20.0 percent of an examined control, a margin of 34.8 points where the preregistered rule asks for 25, and a post-hoc design-weighted estimate places a defective reference query on 17.5 percent of the 508 questions.

The suite also accepts answers that are wrong, and relabelling only the rejected side with the full census reverses the sign of the understatement for every checkpoint (\cref{fig:hero}b). Expert labels on both sides, on XiYan-32B and Kwai-32B, fix it as positive under every convention: those experts call wrong 17.7 and 15.1 percent of the answers the suite accepts, against 36.8 and 36.3 percent of those it rejects. Neither oracle reports the level those labels give. A certificate calibrated at a nominal 0.10 carries 20.0 points of expert-judged risk on XiYan-32B and 17.2 on Kwai-32B; the shipped database understates that on both checkpoints, and the suite understates it on XiYan-32B while overstating it on Kwai-32B (\cref{fig:hero}a). The multi-instance cell carries less risk under every yardstick, and it gets there by changing which questions are answered: the two partitions return the same SQL on 97.1 percent of answers.

The oracle also decides how a confidence score is judged. Each execution-consistency score looks better under the labels of the oracle that built its clusters, in 16 of 16 score-by-checkpoint combinations, while likelihood scores that read no oracle never do. Under suite labels the score built on the suite partition leads by about nine points of AUROC on both audited checkpoints; under expert labels that lead holds on Kwai-32B, 6.96 points at $p = 0.004$, and on XiYan-32B the expert interval, $-1.2$ to 4.3 points, excludes the 8.3 points the suite labels report. Either oracle agrees with the experts on only 74 to 80 percent of the answers the two experts agree on. An oracle-internal gain therefore does not by itself establish a gain under an independent yardstick.

\paragraph{Contributions.}
\begin{enumerate}
    \item We show that the risk a conformal abstention certificate for text-to-SQL reports is set by the execution oracle that labels its calibration set. In a preregistered intervention on Spider-Realistic with four SQL-specialist checkpoints, swapping the benchmark's shipped database for its distilled multi-instance test suite raises the certificate's held-out risk 2.73 to 10.23 points above the risk its own labels report (\cref{sec:oracleswap}).
    \item We audit both oracles against labels from outside them. An AI-labelled census of every answer the stricter oracle rejects attributes most rejections to the benchmark rather than to the system, and a preregistered blinded expert audit supports its reference-defect flag. Blinded expert labels on both sides of the oracle show, post hoc, that a certificate calibrated at a nominal 0.10 carries 20.0 and 17.2 points of risk on two checkpoints, a level neither oracle reports (\cref{sec:census,sec:reshapes,sec:experts}).
    \item We show that the oracle also decides how an execution-consistency score is judged, and test this against an independent yardstick. Each such score looks better under the labels of the oracle that built its clusters, in 16 of 16 combinations, and in a preregistered blinded expert audit the AUROC gain that suite labels report for building the score on suite clusters holds on one checkpoint and falls outside the expert interval on the other (\cref{sec:alignment,sec:yardstick}).
\end{enumerate}

\section{Related work}
\label{sec:related}

\paragraph{Execution-based evaluation and its oracles.}
Since the distilled test suites of \citet{zhong2020testsuite}, work on the oracle has both quantified the failures of single-database execution and proposed evaluation-side alternatives. On the first, \citet{kim2024flex} quantified the false positives and false negatives of execution matching against expert judgement, and \citet{ascoli2024etm} did so under semantic analysis. On the second, \citet{klopfenstein2025spotit} used formal verification to search for a database that separates two queries the shipped database cannot, and \citet{habibollah2026synsql} synthesised databases for the same purpose. \citet{tremante2026spotitplus} mined schema constraints so that the separating database a verifier returns is one a real deployment could hold, which is the concern that our synthetic-instance-defect label of \cref{sec:protocol} records case by case. Separately, and as a candidate-selection method rather than an evaluation-side tool, \citet{li2026dpc} synthesised a minimal distinguishing database at inference time to select among generated SQL candidates. Two of these works already examine what a stricter oracle rejects: \citet{klopfenstein2025spotit} manually inspected the counterexamples its verifier found for a sample of fifty generated queries and attributed each difference to an ambiguous question, an incorrect reference query or an incorrect generated query, reporting the reference at fault more often than the model, and \citet{kim2024flex} decomposed the false negatives of execution matching against expert judgement. We carry that observation from fifty inspected queries to a census of every rejected answer of four checkpoints under five labels, and test the label that carries the most weight in a preregistered blinded human audit. The test suites have also served as a calibration label: \citet{stengel2023calibrated} found a T5-base text-to-SQL parser on the Spider test set far less miscalibrated against test-suite accuracy than against exact match. None of this work has been carried into the layer that certifies a system's risk, which is where our intervention acts. Nor does any of it ask whether a confidence score built on one oracle can be evaluated under that oracle at all.

\paragraph{Benchmark validity, reference-query quality and label noise.}
Annotation errors in text-to-SQL benchmarks are documented. \citet{pourreza2023evaluating} re-evaluated Spider-family predictions by hand and analysed BIRD systematically, finding that underspecified questions and defective reference queries account for a large share of the disagreements between models and benchmarks. \citet{wretblad2024noise} examined noise in BIRD and found reference queries and questions that experts would not accept, and \citet{jin2026annotation} corrected annotation errors with experts and measured their effect on leaderboard rankings. Spider's original exact-match metric did not evaluate literal values \citep{yu2018spider}, so a wrongly cased literal in a reference query was invisible to it on a case-sensitive engine. Outside text-to-SQL, \citet{northcutt2021labelerrors} showed that label errors in test sets are pervasive and that modest increases in their prevalence would reverse model selection. Conformal prediction has its own literature on what a certificate means when the calibration labels are noisy: \citet{einbinder2024labelnoise} showed that under dispersive noise the certificate remains valid but conservative, and \citet{sesia2023noisy} adapted conformal classification to noisy labels.  This paper carries oracle validity into the certificate layer: it reports how much of what the stricter oracle rejects is attributable to the reference queries rather than to the system, validates the suspected-defect label with human experts under a preregistered protocol, and measures what that validity does to the risk a certificate reports.

\paragraph{Uncertainty, calibration and abstention for text-to-SQL.}
Confidence estimation for text-to-SQL has been studied with black-box consistency signals, white-box token probabilities and execution-grounded features \citep{maleki2025confidence}, and with parser-independent error detectors built on structural features of the question and the query \citep{chen2023errordetection} and entropy-based selective classifiers \citep{somov2025confidence}. Calibration has been applied post hoc, by rescaling sequence probabilities \citep{ramachandran2024calibration} or by sub-clause frequencies \citep{liu2025subclause}, and execution-consistency sampling has been given an adaptive stopping rule \citep{anavi2026stop}. \citet{richardson2026predicts} compared these signals as selective predictors and computed a distribution-free selective-risk certificate that triggers only at the loosest risk target on their generator; \citet{ziletti2026ambiguity} separated ambiguity from instability in a clinical setting; \citet{tritto2026verification} verified candidates with outcome reward models. Abstention with guarantees reached text-to-SQL through conformal branch prediction at schema linking \citep{chen2025adaptive} and, end to end, through execution entropy with a conformal layer on top \citep{chen2026endtoend}; \citet{lee2024trustsql} scored reliability with penalties for wrong answers, and \citet{wu2026never} argued for structural abstention where answers are consumed as fact. None of them labels correctness with a multi-instance oracle, and those that execute queries do so on a single database per schema. Our system under test is deliberately the simplest member of this family, self-consistency mass with conformal risk control, so that the oracle rather than the method is the variable, and the five other scores we test are the black-box members of the family under one certificate.

\paragraph{Conformal abstention for language models.}
Conformal risk control extends split conformal prediction to expected control of monotone losses \citep{angelopoulos2024crc}, and the learn-then-test framework gives high-probability control of more general risks \citep{angelopoulos2021ltt,angelopoulos2023gentle}. Conformal abstention has been developed for language models outside text-to-SQL: \citet{abbasiyadkori2024conformal} calibrated an abstention rule on a similarity-based self-consistency score with a distribution-free bound on the hallucination rate, \citet{quach2024conformal} calibrated the sampling and filtering of candidate generations so that the returned set contains an acceptable one with a guaranteed probability, and \citet{mohri2024conformal} filtered the claims of a generation so that what remains is factual with a guaranteed probability. On the limits of such rules, \citet{kotte2026crc} showed that, when the base risk $\mu$ of a generator exceeds $\alpha$, any distribution-free method must abstain on a fraction of at least $(\mu - \alpha)/(1 - \alpha)$ of inputs, \citet{xu2026geometry} calibrated abstention geometrically, and \citet{marques2025coverage} gave the law of the empirical coverage of split conformal prediction, which we use to read resplit variability. The methods of \citet{abbasiyadkori2024conformal}, \citet{quach2024conformal} and \citet{mohri2024conformal} take their correctness labels from a reference or a judge and treat them as given; what the certificate carries when that label comes from a lenient execution oracle is what \cref{sec:oracleswap} measures.

\paragraph{Black-box confidence scores and correctness functions.}
The scores we compare come from the general literature on black-box uncertainty for language models: self-consistency mass \citep{wang2023selfconsistency}, discrete semantic entropy \citep{kuhn2023semantic,farquhar2024hallucinations}, the degree measure of \citet{lin2024generating} and the set count \citep{kuhn2023semantic,lin2024generating}, and length-normalised sequence log-probability \citep{malinin2021uncertainty}; implementations of these score families are available in UQLM \citep{bouchard2025uqlm} and LM-Polygraph \citep{fadeeva2023polygraph}, and functional entropy transfers the semantic-cluster idea to code without executing it \citep{bouchard2026functional}.  That the choice of correctness function changes AUROC and the ranking of uncertainty methods is known outside text-to-SQL: \citet{ielanskyi2026pitfalls} showed that approximate correctness functions disagree with one another enough to reorder methods and to inflate a method's apparent performance. What is specific here is that the execution oracle does not only supply the labels; it also builds the clusters the four consistency scores are read from, so the label and the score are not independent; \cref{sec:alignment} measures that coupling as a signed oracle-by-partition interaction and \cref{sec:yardstick} tests it against labels neither oracle produced.

\section{Experimental design}
\label{sec:design}

\begin{figure}[t]
    \centering
    \input{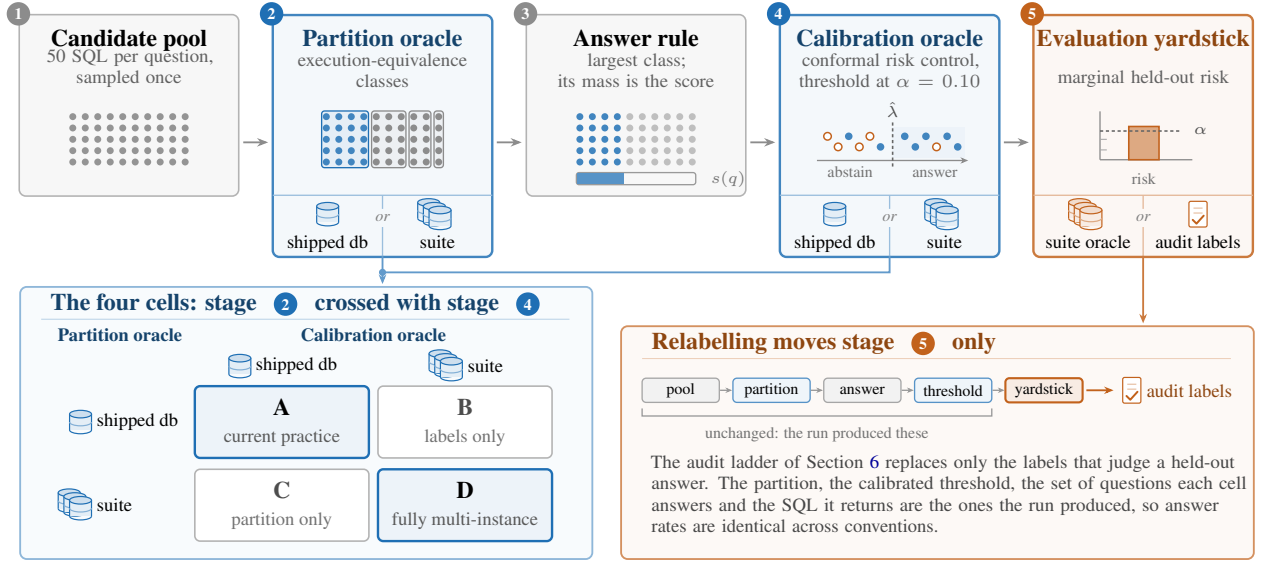}
    \caption{Where the oracle enters the certificate. A question's fifty sampled candidates (1) are cut into execution-equivalence classes (2); the system answers with the representative of the largest class and scores the question by that class's mass (3); conformal risk control picks a threshold on that score at the nominal level (4); and the system's marginal held-out risk, with abstentions counted as zero loss, is read off a yardstick (5). In stage 4 each dot is a calibration question placed at its score, drawn hollow when its answer is wrong. Stages 2 and 4 consult an oracle, that is, the databases a query is executed on, and each can take the single database the benchmark ships (one cylinder) or every instance of the distilled suite (stacked cylinders); crossing the two choices gives the four cells of \cref{sec:cells}, cell A being current practice and cell D fully multi-instance. The candidate pool, the answer rule and the threshold procedure are held fixed across every cell. The panel at lower right marks where the audit ladder of \cref{sec:reshapes} acts, on the labels of stage 5 alone.}
    \label{fig:pipeline}
\end{figure}

This section fixes everything the intervention holds constant. \Cref{sec:system} defines the system under test, a self-consistency score over execution-equivalence classes with a split-half conformal certificate on top, and \cref{sec:oracles} the two oracles that can supply those classes and the correctness labels. \Cref{sec:cells} crosses the two choices into four cells and names the two quantities this paper reports; \cref{fig:pipeline} shows where in the pipeline each choice acts. \Cref{sec:data,sec:prereg_rules} give the pools and the rules frozen before any of them was generated, \cref{sec:protocol} the protocol that labels what the stricter oracle rejects, and \cref{sec:designchoices} the analysis choices that were declared rather than preregistered.

\subsection{The system under test}
\label{sec:system}

For a question $q$ over a schema, the generator draws $m = 50$ candidate SQL queries $S_q$ independently at temperature 1.0 and top-$p$ 0.95. An execution oracle $O$ maps a query to a result signature on a set of database instances; two candidates are execution-equivalent under $O$ if the official test-suite comparator of \citet{zhong2020testsuite} judges their results equal on every instance of $O$, with the row-order flag set when the reference query of $q$ carries an \texttt{ORDER BY}. A union-find pass over the pairwise comparisons partitions the usable candidates, those that parse and were not truncated, into execution-equivalence classes $\Pi_O(S_q)$. The system answers with the representative of the largest class, and its confidence score is the mass of that class,
\[
  \score_O(q) = \frac{\max_{C \in \Pi_O(S_q)} |C|}{\lvert \text{usable candidates of } q \rvert},
\]
the self-consistency score of \citet{wang2023selfconsistency} on execution classes. The preregistered construction is an offline benchmark analysis: the reference query takes part in the union-find, and the row-order flag is read from it. A deployed system has no reference query, so the operational definition of the score is the gold-free construction: the partition is built from the sampled candidates alone, the row-order flag is set when either query of the compared pair carries an \texttt{ORDER BY}, and the reference query is used only to label a class after the partition exists. \Cref{sec:robustness} reports what the preregistered and the gold-free constructions differ by, together with an intermediate construction that isolates the cost of reading the flag per pair. A class is correct under an oracle $O'$ if its representative agrees with the reference query on every instance of $O'$. The certificate is split-half conformal risk control \citep{angelopoulos2024crc}: on a calibration half of the questions, with loss one when the system answers and its class is wrong under the calibration oracle and zero when it abstains, the threshold is
\[
  \hat\thr = \inf\Big\{\thr \in G : \tfrac{n}{n+1}\,\hat R_n(\thr) + \tfrac{1}{n+1} \le \alpha\Big\},
\]
where $\hat R_n(\thr)$ is the empirical loss of answering exactly the calibration questions with $\score_O(q) \ge \thr$, $G$ is a grid of 201 thresholds on $[0, 1]$, and $\alpha = 0.10$; if no threshold satisfies the bound, the system abstains on everything. On the held-out half the system answers the questions with $\score_O(q) \ge \hat\thr$, and we record its held-out empirical marginal risk, the fraction of all held-out questions on which it answers and is wrong under the evaluation oracle; its answer rate; and secondarily its selective risk, the fraction wrong among answered questions, which selective classification in the sense of \citet{geifman2017selective} controls and the certificate does not. Each pool is analysed over 200 random split-halves under two schemes. Question splits draw the calibration half from the questions and are the setting in which the conformal guarantee applies. Database splits calibrate on 9 of the 19 schemas and evaluate on the other 10, so that no schema appears on both sides; the calibration questions of 9 schemas are not exchangeable with the questions of 10 unseen schemas, so database splits are an empirical schema-disjoint stress test of the same thresholds and not a distribution-free certificate for unseen schemas.

\subsection{The two oracles}
\label{sec:oracles}

The weak oracle is the single database the benchmark ships with each schema, the oracle of the text-to-SQL uncertainty pipelines of \cref{sec:related} that read confidence off execution consistency. The multi-instance suite oracle is the distilled test suite of \citet{zhong2020testsuite}: for the 19 schemas of Spider-Realistic, 651 distilled instances, between 24 and 59 per schema, plus the shipped database, 670 database files in all, on every one of which two queries must agree. The suite label refines the weak label: a class the suite accepts is accepted by the shipped database, and we verified that no class in the study is accepted by the suite and rejected by the shipped database. Two pilot measurements on official Spider dev, 1,034 questions over 20 schemas, motivated the intervention. First, a collision census: for every pair of different questions on the same schema whose reference queries return identical order-conditional canonical results on the shipped database, 688 pairs, we re-executed both on every suite instance; 3 pairs agree everywhere, so 0.44 percent of the collisions survive the suite. Second, a near-miss census: we applied single-edit mutations to reference queries (dropping a conjunct, swapping an aggregate, flipping a comparison, dropping \texttt{DISTINCT}, changing the ordering direction and similar edits), executed the 856 mutants on every instance, and kept the 832 whose result differs from the original on at least one suite instance; under an order-conditional canonical comparison of results, the shipped database alone accepts 7.0 percent of those as equivalent to the original. Both are computed from reference queries, the first on pairs of them and the second on their mutants; the intervention measures the model's candidates.

\subsection{The intervention and its two quantities}
\label{sec:cells}

The partition that defines the score and the labels that calibrate the threshold can each come from the weak oracle or from the suite, giving four cells. Cell $\cellA$, weak partition and weak labels, is current practice. Cell $\cellB$ keeps the weak partition and calibrates on suite labels; cell $\cellC$ builds the partition from the suite and calibrates on weak labels; cell $\cellD$ is fully multi-instance. Every cell is evaluated under the suite oracle. Two quantities summarise the intervention, both in percentage points of held-out empirical marginal risk:
\[
  \GAP = \risk^{\suite}(\cellA) - \risk^{\weak}(\cellA), \qquad
  \DmA = \risk^{\suite}(\cellD) - \risk^{\suite}(\cellA),
\]
where $\risk^{O}(X)$ is the held-out empirical marginal risk of cell $X$ under oracle $O$ at the threshold its calibration selected, averaged over splits and seeds. The estimand is descriptive: the three-seed mean over 200 split-halves of this fixed benchmark and these pools. The 200 splits overlap, so the sign shares reported with every contrast describe resplit stability and are not confidence levels, and no inference to other questions or schemas is claimed beyond the schema-disjoint stress test of the database splits. The certificate itself promises only the nominal bound $\alpha$; $\risk^{\weak}(\cellA)$ is what a practitioner who evaluates on the shipped database measures, and $\GAP$ is the amount by which that measurement understates the risk under the suite oracle. It is non-negative by construction because the suite label refines the weak label. $\DmA$ is the change in suite-oracle risk from moving to the fully multi-instance cell, negative when cell $\cellD$ carries less.

\subsection{Data, checkpoints and pools}
\label{sec:data}

The main data are all 508 questions of Spider-Realistic \citep{deng2021spiderrealistic}, the paraphrased subset of Spider dev \citep{yu2018spider} over 19 schemas; a matched control uses the one-to-one paired original questions. The checkpoint panel (\cref{tab:panel}) was chosen by a frozen entry rule on a disjoint pilot of 200 Spider dev questions: parse rate at least 0.98, truncation rate at most 0.01, and top-1 accuracy under the suite oracle at least 0.75. Four SQL-specialist checkpoints entered, from two Qwen lineages: Kwai-AutoSQL-32B and Kwai-AutoSQL-14B, fine-tuned from Qwen3 \citep{kwai2026autosql32b,kwai2026autosql14b}, and XiYanSQL-QwenCoder-32B-2504 \citep{liu2025xiyansql}, from the Qwen2.5 lineage, and OmniSQL-32B \citep{li2025omnisql}, fine-tuned from Qwen2.5-Coder. SQLCoder-70B-alpha was excluded at 0.67 on the pilot, and llama-3-sqlcoder-8b produced no parseable output under two inference back-ends; the Meta lineage has no member in the panel. Each checkpoint has three generation seeds, 101, 202 and 303, giving twelve pools of 508 questions and 50 samples, 304,800 generations in all, plus one false-schema control pool per checkpoint in which each question is paired with the schema of an unrelated database. Parse rates are at least 0.9985 and no generation was truncated in any main pool.

\begin{table}[t]
\centering
\caption{The checkpoint panel. Entry was decided by a frozen rule on a disjoint 200-question pilot of Spider dev (parse rate at least 0.98, truncation rate at most 0.01, top-1 accuracy under the multi-instance suite oracle at least 0.75). Main pools are 508 Spider-Realistic questions with 50 samples each at temperature 1.0 and top-$p$ 0.95, three generation seeds per checkpoint; main-pool top-1 is the range over seeds, parse rate the minimum over seeds and truncation rate the maximum. The short names in parentheses are used in the other tables and figures. The panel spans two Qwen lineages, Qwen3 and Qwen2.5. SQLCoder-70B-alpha (0.67 on the pilot) and llama-3-sqlcoder-8b (no parseable output) were excluded.}
\label{tab:panel}
\small
\setlength{\tabcolsep}{4pt}
\begin{tabular}{p{0.34\linewidth}lcccc}
\toprule
Checkpoint & Base model & Pilot top-1 & Main-pool top-1 & Parse rate & Truncation \\
\midrule
Kwai-AutoSQL-32B (Kwai-32B) & Qwen3-32B & 0.825 & 0.738--0.740 & 0.9985 & 0.0000 \\
Kwai-AutoSQL-14B (Kwai-14B) & Qwen3-14B & 0.830 & 0.744--0.754 & 0.9998 & 0.0000 \\
OmniSQL-32B (Omni-32B) & Qwen2.5-Coder-32B & 0.770 & 0.728--0.730 & 0.9992 & 0.0000 \\
XiYanSQL-QwenCoder-32B-2504 (XiYan-32B) & Qwen2.5 family & 0.925 & 0.860--0.862 & 1.0000 & 0.0000 \\
\bottomrule
\end{tabular}
\end{table}

\subsection{Preregistration and analysis rules}
\label{sec:prereg_rules}

The panel, the seeds, the two split schemes, the entry rule and the aggregation rule were frozen before any pool of the panel was generated, in the third of three preregistrations; the first two, on one checkpoint, are described in \cref{app:history}. The frozen rule declares $\GAP$ positive for a checkpoint if every seed is above zero and the three-seed mean is at least 0.010 under both split schemes, and declares the $\DmA$ contrast passed if every seed is below zero and the mean is at most $-0.030$ under both schemes, robustly passed if every seed is at most $-0.030$; a lineage is judged by its highest-ranked member. The registration's outcome table names the results: O1 if at least two lineages have positive $\GAP$ and at least two pass the $\DmA$ bar with one robust pass, O2 if at least two lineages have positive $\GAP$ but fewer than two pass the bar or none passes robustly, O4 if no new lineage has positive $\GAP$, and O5 if no candidate enters. The realised outcome is O2 under the frozen code. The instrument is checked at every tier by parse rate, truncation rate and the false-schema control, whose preregistered expectation is a suite top-1 accuracy at most 0.10 and a cell-$\cellA$ suite risk at most $\alpha + 0.02$. Apart from the analyses that the two expert audits preregistered (\cref{app:humanaudit,app:yardstick}), all analyses beyond the preregistered result, including the alpha curves, the frontier, the exhaustive database splits, the cross-fit, the gold-free construction, the six-score comparison, both censuses, both audit ladders, the decomposition of the $\DmA$ contrast and the oracle-alignment finding, are post hoc and are labelled so where they appear. \Cref{app:repro} says what is released with this paper and what is not, and gives the address of the archive.

\subsection{The census protocol}
\label{sec:protocol}

To learn what the suite's rejections are, we labelled, post hoc, every answer it rejects in two populations. For each distinct case, a question together with the returned SQL, a mechanical step first re-executed the reference query and the returned query on the shipped database and on every suite instance with the same comparator and row-order convention as the analysis, and recorded the first instance on which they disagree, the two result tables truncated to twelve rows, a difference signature (row order only, duplicate multiplicity only, different column count, one result a subset of the other, partial overlap, disjoint, one side empty), and surface features of both queries.  A dossier then presented each case with its question, both queries, the disagreement breadth and both tables, followed by the full schema, so that a labeller could judge without opening the benchmark.

Labels were assigned by an AI reviewer, Codex running gpt-5.6-sol, in two independent passes that read the same dossier and could not see each other: the second pass re-batched the cases so that no case kept its neighbours, and listed the taxonomy in reverse order, to counter within-batch anchoring and list-order preference, the latter a documented bias of language-model evaluators \citep{zheng2023judge,wang2024fair}. On the full census the second pass was also told that 472 of the 742 cases already disagree on the shipped database, where a synthetic-instance-defect label would call the benchmark's own database defective (\cref{app:census}). A third fresh thread adjudicated every disagreement with both labels and both rationales in view. Every label came from these passes and the adjudicating thread. The taxonomy has five labels: semantic error (the returned query answers the question wrongly), underspecified question (the question does not fix the point on which the two queries differ), suspected reference-query defect (the reference is the worse rendering of the question), synthetic-instance defect (the two queries differ only on values a real database of that schema would not contain), and comparator artefact (same information, judged unequal). For semantic errors an error type was also recorded. The labeller is one AI reviewer family and is not the Qwen family of the tested generators; of the five labels, human SQL experts have tested only the suspected reference-query defect, in the preregistered blinded audit of \cref{app:humanaudit}. The agreement statistics of \cref{sec:agreement} measure the reproducibility of the labelling, not its correctness.

\subsection{Declared design choices}
\label{sec:designchoices}

Three choices affect how the numbers should be read. First, the score divides by the number of usable candidates, a statistic of the model's own output; the alternative budget denominator, which divides by all 50 samples and treats unusable ones as implicit abstention mass, changes the question-split $\GAP$ and $\DmA$ by at most 0.14 points on the main pools, where at least 99.85 percent of candidates are usable. Second, 53 of 12,192 top-class decisions are tied, and the frozen analyser and the per-question files it archived break those ties differently; every recomputation from those files, including both censuses, the audit ladders and the six-score comparison, follows the file order, and recomputing every reported $\GAP$ and $\DmA$ under the frozen rule moves no value by more than 0.085 points and changes no sign (\cref{app:tie}). Third, all four cells share one answer rule, the representative of the largest class, so the six-score comparison compares six abstention rules of one system, not six end-to-end methods; and the four consistency scores share the equivalence relation under test, so only the two likelihood scores are independent of it.

\section{The oracle swap changes what the certificate reports}
\label{sec:oracleswap}

This section reports the preregistered measurement, with its post-hoc robustness checks. The multi-instance suite oracle is the yardstick throughout, and risk is held-out empirical risk under the named oracle at the threshold each cell's calibration selected. 

\subsection{Preregistered result}
\label{sec:prereg}

\Cref{tab:main} gives the preregistered result. For every checkpoint, every generation seed and both split schemes, the current-practice cell $\cellA$ carries more risk under the suite oracle than under its own weak labels, and the fully multi-instance cell $\cellD$ carries less suite-oracle risk than cell $\cellA$. $\GAP$ under question splits is 10.23, 8.08, 9.51 and 3.11 points for Kwai-32B, Kwai-14B, Omni-32B and XiYan-32B, and 10.17, 7.86, 9.56 and 2.73 points under database splits; $\DmA$ is $-10.38$, $-8.30$, $-9.59$ and $-2.90$ points under question splits and $-10.70$, $-8.02$, $-10.30$ and $-1.26$ under database splits. The preregistered range of $\GAP$ over the eight checkpoint-and-split combinations is therefore 2.73 to 10.23 points. Generation seeds move these values little: over both split schemes, per-seed $\GAP$ spans 9.90 to 10.41 points for Kwai-32B, 7.51 to 8.30 for Kwai-14B, 9.44 to 9.63 for Omni-32B and 2.62 to 3.13 for XiYan-32B, and per-seed $\DmA$ spans $-10.02$ to $-10.98$, $-7.88$ to $-8.47$, $-9.41$ to $-10.45$ and $-1.11$ to $-2.93$ points. Under the frozen rule, $\GAP$ is positive for all four checkpoints in both lineages, and the $\DmA$ bar of three points is passed robustly by Kwai-32B, Kwai-14B and Omni-32B and missed by XiYan-32B, by 0.10 points under question splits and by 1.74 points under database splits. The preregistered outcome is therefore O2: the measurement claim holds across the two lineages; the contrast claim does not constitute a cross-lineage result. The contrast comes with a change in answer rate: cell $\cellD$ answers 13.9 to 17.7 points fewer questions than cell $\cellA$ for three checkpoints, and 4.9 points fewer, or 2.8 points more under database splits, for XiYan-32B. Across resplits, $\DmA$ is negative in 99.7 to 100 percent of the 200 splits for three checkpoints, on average over seeds, and in 66.5 to 72.5 percent per seed for XiYan-32B under database splits. \Cref{fig:cells} in \cref{app:more} shows all four cells.

\begin{table}[t]
\centering
\caption{Preregistered main result at nominal risk $\alpha=0.10$, three-seed means. The current-practice cell A is calibrated on labels from the shipped database and evaluated under the multi-instance suite oracle; GAP is its suite-oracle risk minus the risk it reports on its own labels; D minus A is the suite-oracle risk of the fully multi-instance cell D minus that of cell A, negative meaning D carries less. Answer-rate change is D minus A. The last two columns are the frozen-rule verdicts (GAP positive: every seed above zero and mean at least 0.010; D minus A pass: every seed below zero and mean at most $-0.030$; robust: every seed at most $-0.030$). All risks are under the suite oracle unless marked reported.}
\label{tab:main}
\footnotesize
\setlength{\tabcolsep}{3.5pt}
\begin{tabular}{llcccccccc}
\toprule
Checkpoint & Split & A reported & A suite & GAP (pts) & D suite & $\DmA$ (pts) & $\Delta$answer (pts) & GAP$>0$ & $\DmA$ rule \\
\midrule
Kwai-32B & question & 0.0981 & 0.2004 & +10.23 & 0.0966 & -10.38 & -17.0 & yes & pass (robust) \\
 & database & 0.1042 & 0.2059 & +10.17 & 0.0989 & -10.70 & -17.7 & yes & pass (robust) \\
Kwai-14B & question & 0.0997 & 0.1805 & +8.08 & 0.0976 & -8.30 & -13.9 & yes & pass (robust) \\
 & database & 0.0970 & 0.1756 & +7.86 & 0.0954 & -8.02 & -14.0 & yes & pass (robust) \\
Omni-32B & question & 0.0948 & 0.1899 & +9.51 & 0.0941 & -9.59 & -14.1 & yes & pass (robust) \\
 & database & 0.1096 & 0.2051 & +9.56 & 0.1021 & -10.30 & -14.5 & yes & pass (robust) \\
XiYan-32B & question & 0.0984 & 0.1295 & +3.11 & 0.1005 & -2.90 & -4.9 & yes & fail \\
 & database & 0.0892 & 0.1165 & +2.73 & 0.1039 & -1.26 & +2.8 & yes & fail \\
\bottomrule
\end{tabular}
\end{table}

\subsection{Across the nominal level}
\label{sec:alpha}

The understatement is not a property of one nominal level. When the certificate is recomputed post hoc from the per-question sufficient statistics at $\alpha \in \{0.05, 0.10, 0.15, 0.20\}$ (\cref{fig:alpha}, appendix), the suite-oracle risk of cell $\cellA$ is 2.1 to 2.6 times the nominal level at $\alpha = 0.05$, 1.8 to 2.0 times at $0.10$ and 1.6 to 1.7 times at $0.15$ for the three checkpoints other than XiYan-32B, and the ratio falls to 1.3 to 1.4 at $\alpha = 0.20$ because by then cell $\cellA$ answers every question. For the same three checkpoints under both split schemes, cell $\cellD$ sits at 0.039 to 0.051, 0.094 to 0.102 and 0.144 to 0.153 at the three lower levels. The current-practice cell of XiYan-32B saturates in both directions under question splits: at $\alpha = 0.05$ it answers 5.0 percent of questions and carries 0.006 suite-oracle risk, and from $\alpha = 0.15$ it answers everything.

\subsection{Labels or partition}
\label{sec:frontier}

The intervention changes two things at once, the labels the certificate is calibrated on and the partition that defines its score, and cells $\cellB$ and $\cellC$ separate them. At matched nominal level, cell $\cellB$ (suite labels on the original partition) is the more conservative of the two, answering 46.6 to 74.0 percent of questions against 74.1 to 92.3 percent for cell $\cellD$ at $\alpha = 0.10$ under question splits, while the two realised risks differ (0.072 to 0.092 against 0.094 to 0.101), so the comparison has to be made at a common risk ceiling. With every distinct top-class mass and the abstain-on-everything point enumerated (\cref{fig:frontier}, appendix), at an empirical ceiling of 0.05 cell $\cellB$ can only abstain on everything for all four checkpoints while cell $\cellD$ answers 58.7, 55.6 and 63.3 percent for Kwai-32B, Kwai-14B and Omni-32B; at 0.075 three checkpoints still cannot answer under the label-only cell while the suite-partition cell answers 67.3, 70.3 and 87.9 percent, Omni-32B being the exception whose label-only cell answers 54.5 percent; at a ceiling of 0.10 the difference closes to 0.4 to 8.9 points, and at 0.125 and 0.15 it closes to 0.0 to 4.1 points. This is an empirical frontier on the full data under a common ceiling rather than a common realised risk, and it is a post-hoc description: within this setting, rebuilding the equivalence classes with the suite is what moves the frontier, and relabelling alone does not.

\subsection{Six scores, one certificate}
\label{sec:sixscores}

\begin{figure}[t]
    \centering
    \includegraphics[width=\textwidth]{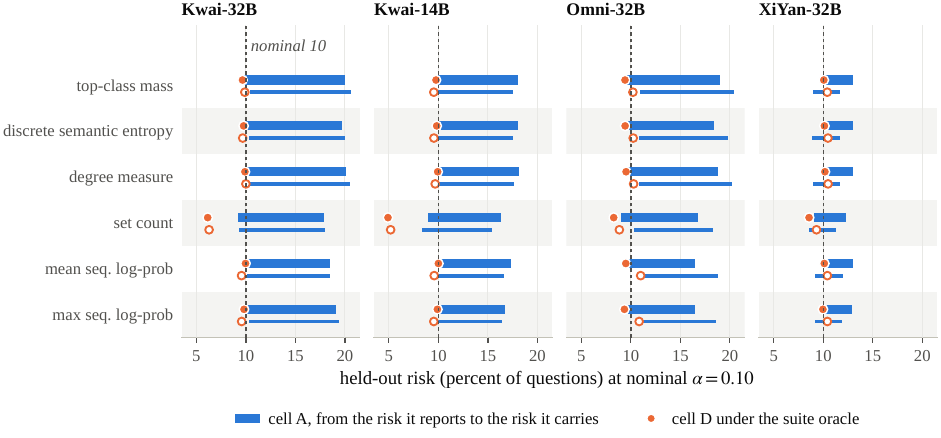}
    \caption{What the certificate reports and what it carries, at nominal risk 0.10 under the preregistered suite labels, for six confidence scores, four checkpoints and two split schemes, with one certificate, one answer rule, the same pools and the same splits. Each bar starts at the risk the current-practice cell $\cellA$ reports on its own labels, within 1.7 points of the nominal 10 in every cell, and ends at the risk it carries under the suite oracle, so the length of the bar is $\GAP$; the circle is where the fully multi-instance cell $\cellD$ lands under the same oracle, and its distance back from the end of the bar is D minus A. The thick upper mark of each row carries the question-split value and the thin lower mark the database-split value. All 48 bars run to the right and every circle falls short of the end of its bar, whichever score sets the threshold.}
    \label{fig:sixscores}
\end{figure}

Is the understatement a property of top-class mass? In a post-hoc extension, holding the pools, the questions, the 200 splits, the certificate and the answer rule fixed and changing only the score that decides whether to answer, we repeated the intervention for discrete semantic entropy, the degree measure of \citet{lin2024generating}, the set count \citep{kuhn2023semantic,lin2024generating}, and mean and maximum length-normalised sequence log-probability. Every score is a confidence score, with larger values meaning greater confidence, and the certificate answers when the score is at least the calibrated threshold. With $p_C = |C| / u_q$ the mass of class $C$ among the $u_q$ usable candidates of $q$: top-class mass is $\max_C p_C$; discrete semantic entropy enters as $\sum_C p_C \log p_C$, the negative entropy of the class masses; the degree measure is $\sum_C p_C^2$; the set-count measure enters as minus the number of classes; and the two likelihood scores are the mean and the maximum over usable candidates of the candidate's cumulative token log-probability divided by its token count, which read no oracle. Unusable candidates are removed before any score is computed. Each score has a fixed threshold grid that is never adapted to a split: 201 equally spaced points on $[0, 1]$ for top-class mass and the degree measure, 201 points on $[-\log 50, 0]$ for the entropy score, the integers $-50$ to $-1$ for the set count, and steps of $0.001$ on $[-8, 0]$ together with $-\infty$ for the two likelihood scores; refining every grid five-fold changes $\GAP$ and $\DmA$ by at most 0.13 points (\cref{app:instrument}). \Cref{fig:sixscores} shows the 48 $\GAP$ and 48 $\DmA$ values and \cref{tab:sixscores} (appendix) lists them. Across the 48 score-by-checkpoint-by-split cells $\GAP$ lies between 2.71 and 10.23 points, and the mean $\DmA$ is negative in all 48, between $-1.19$ and $-11.78$ points; the cells are not independent replications, because they share the pools, the questions and the splits. The operating points are equally uniform: at $\alpha = 0.10$ the current-practice cell answers 72 to 98 percent of questions and carries 1.13 to 2.06 times the nominal level under the suite oracle across the 48 cells, and the spread across scores is smaller than the spread across checkpoints. Among the six scores tested, changing the scoring function does not rescue the certificate from the oracle.

Two of the six scores never read any oracle, and they understate as much as the four that do. This settles a sufficiency question: coupling between the score and the oracle is not necessary for the understatement, because a score that cannot see the oracle carries it too. A score-free reference quantity makes the point sharper. Call a question a flip question if the answer that cell $\cellA$ returns is accepted by the shipped database and rejected by the suite; the flip rate is 10.96, 8.86, 11.02 and 3.35 percent of questions for Kwai-32B, Kwai-14B, Omni-32B and XiYan-32B. On each split, a certificate's $\GAP$ is the number of held-out flip questions it answers as a share of all held-out questions, so a certificate that never abstains carries the held-out flip rate as its $\GAP$. Under question splits, abstention leaves 60.8 to 95.0 percent of the flip questions answered across the six scores, and the enrichment of flip questions among answered questions relative to their base rate is 84.0 to 102.5 percent: no tested score removes most of the disagreement, and for five score-and-checkpoint combinations the answered questions are slightly richer in flip questions than the pool. This does not decompose $\GAP$ into a label part and a score part, because $\GAP$ is a joint quantity of the labels and the answered set and different scores answer different questions.

\subsection{Robustness of the measurement}
\label{sec:robustness}

Six post-hoc checks, tabulated in \cref{tab:robustness} (appendix), bound the ways the measurement could be an artefact. Instance cross-fit builds the classes and the calibration labels from one half of each schema's suite instances and judges correctness on the other half; in all 24 runs $\GAP$ stays positive and $\DmA$ negative, and the held-out suite risk of cell $\cellD$ lands at 0.094 to 0.101, so the contrast holds on suite instances the calibration did not see. The same check shows how redundant the suite is for this purpose: the two halves disagree on the label of about 0.6 percent of single-instance classes, 1.9 percent of multi-instance classes and 1.6 percent of top classes, which is why cross-fitting cannot stand in for semantic evidence. The candidates-only partition, which keeps the reference query out of the union-find and uses it only to label a class afterwards, and the gold-free construction, which additionally takes the row-order convention from the compared pair rather than from the reference query, reproduce the question-split $\GAP$ and $\DmA$ of every checkpoint to within 0.03 points (\cref{app:instrument}). Over all 92,378 schema-disjoint 9/10 splits of the 19 schemas, $\GAP$ is positive in 99.96 to 100 percent of splits and $\DmA$ negative in 99.83 to 100 percent for three checkpoints, and in 88.5 to 92.8 and 70.0 to 75.1 percent for XiYan-32B; leaving one schema out moves $\GAP$ by at most 2.5 points (Omni-32B without world\_1) and by at most 1.9 points for XiYan-32B (without car\_1). Finally, a matched control that replaces every Spider-Realistic question by its one-to-one paired original Spider question, same schemas, reference queries, prompt template and seeds, gives $\GAP$ of 7.48 and 7.25 points and $\DmA$ of $-7.92$ and $-7.24$ points for Kwai-14B under question and database splits, against 8.08, 7.86, $-8.30$ and $-8.02$ on Spider-Realistic: for Kwai-14B the effect is not created by the paraphrased questions. The false-schema negative control (\cref{fig:negctl}, appendix) confirms that the instrument largely abstains when the signal is destroyed: the control cell answers 0 to 11.0 percent of questions.

A seventh check asks whether the understatement is an artefact of how finely the score resolves. $\GAP$ is a quantity of cell $\cellA$ alone, so drawing $m$ of each question's usable candidates without replacement and recounting the shipped-database partition reduces the budget exactly for it. Over $m \in \{10, 20, 30, 50\}$ and both split schemes the three-seed mean $\GAP$ spans 0.41, 0.47, 0.67 and 0.48 points for Kwai-32B, Kwai-14B, Omni-32B and XiYan-32B, while the number of distinct top-class masses the score can take rises from 7 to 9 at $m = 10$ to between 24 and 46 at $m = 50$ and the share of questions placing every candidate in one class falls by 2 to 14 points. The flip rate moves by at most 0.26 points over the same range. The understatement tracks the flip rate, which is a property of the oracle pair and the checkpoint, rather than the resolution of the score, which the budget changes by a factor of three to five.

\section{A census of everything the suite rejects}
\label{sec:census}

The second question is what the stricter oracle's rejections are made of. Every semantic label in this section was assigned by the two-pass adjudicated AI audit of \cref{sec:protocol}.

\subsection{Why these populations carry the whole result}
\label{sec:populations}

$\GAP$ is carried entirely by the answers of cell $\cellA$ that the shipped database accepts and the suite rejects, because the suite label refines the shipped-database label: across the 11,351 execution-equivalence classes of the twelve pools, no class is accepted by the suite and rejected by the shipped database. Those answers are the disagreement census: 521 answer occurrences, 240 distinct question-and-SQL cases, 99 questions, 13 schemas. $\DmA$ additionally involves answers both oracles reject, because they cancel inside $\GAP$ and do not cancel between two cells with different partitions; the full census covers every answer that cell $\cellA$ or cell $\cellD$ returns and the suite rejects: 2,779 occurrences, 742 distinct cases, 203 questions; of those cases, the 240 disagreement cases are carried over with their labels and 502 are labelled for the first time. The two cells return the same SQL on 5,919 of 6,096 answers, 97.1 percent (\cref{sec:triage}).

\subsection{Agreement between the passes}
\label{sec:agreement}

\Cref{tab:agreement} reports how reproducible the labelling is. On the disagreement census the two passes agree on the five-way label for 85.0 percent of cases and on the binary semantic-error-or-not label for 92.9 percent, Cohen's $\kappa$ \citep{cohen1960kappa} of 0.787 and 0.801; on the 502 newly labelled cases of the full census they agree for 91.6 and 96.8 percent, $\kappa$ of 0.873 and 0.935. At least one pass flagged 68.8 percent of the disagreement cases and 44.7 percent of the full-census cases as borderline; 36 and 78 disagreements went to adjudication, and none remained unresolved. Where human experts have tested one of these labels, the suspected reference-query defect, it discriminates between defective and sound reference queries, and expert confirmation tracks the strength of the AI label (\cref{sec:c7}).

\begin{table}[t]
\centering
\caption{The two censuses and the agreement of the two independent labelling passes, computed from the case-level labels. Labels were assigned by the two-pass adjudicated AI audit. Binary agreement collapses the five labels to semantic error versus not. In the four agreement and $\kappa$ rows, the full-census value is over all 742 cases and the parenthesised value is over the 502 cases labelled for the first time in that census, the other 240 being carried over from the disagreement census with their labels.}
\label{tab:agreement}
\small
\begin{tabular}{p{0.36\linewidth}>{\raggedright\arraybackslash}p{0.27\linewidth}>{\raggedright\arraybackslash}p{0.27\linewidth}}
\toprule
 & Disagreement census & Full census \\
\midrule
Population & weak-accepted, suite-rejected A-cell answers & any A- or D-cell answer the suite rejects \\
Answers & 521 & 2,779 \\
Distinct cases & 240 & 742 \\
Questions & 99 & 203 \\
Five-label agreement & 85.0\% & 89.5\% (91.6\%) \\
Binary agreement & 92.9\% & 95.6\% (96.8\%) \\
Cohen's $\kappa$, binary & 0.801 & 0.905 (0.935) \\
Cohen's $\kappa$, five-label & 0.787 & 0.851 (0.873) \\
Cases with a borderline flag in either pass & 165 (68.8\%) & 332 (44.7\%) \\
Adjudicated disagreements & 36 & 78 \\
Unresolved after adjudication & 0 & 0 \\
\bottomrule
\end{tabular}
\end{table}

\subsection{The disagreement census}
\label{sec:c6}

Of the 521 answers the shipped database accepts and the suite rejects, the audit labelled 136, or 26.1 percent, semantic errors (65 of the 240 cases). The rest were labelled underspecified questions (195 answers, 37.4 percent), suspected reference-query defects (107, 20.5 percent) or synthetic-instance defects (83, 15.9 percent); and no case received the comparator-artefact label. The semantic errors concentrate: the 65 cases fall on 30 questions in 7 schemas, of the 13 schemas the disagreement census touches, and are dominated by grouping by name rather than key and by filters attached to the wrong entity. Their share is similar across checkpoints, 22.2 to 29.4 percent of each checkpoint's audited answers.

\subsection{The full census and the suspected reference-query defects}
\label{sec:c7}

In the broader census of 2,779 answers rejected by the suite, representing 742 distinct cases across 203 questions, the audit labelled 972 answers, 35.0 percent, semantic errors (288 cases); on the two checkpoints whose answers two SQL experts judged, the experts call wrong 36.8 and 36.3 percent of the answers the suite rejects (\cref{sec:expertrisk}). The audit assigned the suspected reference-defect label to 213 cases, 887 answers, flagging 73 of the 508 questions as having at least one suspected defective reference query, and the preregistered blinded audit of \cref{app:humanaudit} supports the flag: experts judge 54.8 percent of the flagged questions defective, against 20.0 percent of examined questions without the flag and 7.5 percent of questions on which the suite rejected no answer, a margin of 34.8 points over the examined control where the preregistered rule asks for 25. Thirty-four of the flagged questions carry at least one non-borderline suspected-defect judgement agreed by both passes, twenty-nine carry an agreed judgement that at least one pass marked borderline, and ten depend entirely on adjudication; the experts confirmed 91.2, 27.6 and 10.0 percent of the three groups, so how strongly the AI labelled a question is what predicts whether an expert agrees with it. Weighting the audit's three strata by their sizes, a post-hoc estimate puts a defective reference query on 17.5 percent of the 508 questions, with an interval of 12.2 to 22.8 percent. \Cref{tab:labels} and \cref{fig:defects} in the appendix give the full distributions by case, by answer and by schema; world\_1 alone carries 68 of the 213 cases on 26 questions.

The suspected defects are not scattered; they recur in a handful of families, each mechanically visible once the reference query is executed. The examples below quote the audited cases, with the benchmark's spacing before punctuation removed, and we executed each reference query on the shipped database ourselves. A case-sensitive string literal returns nothing on the database it was written for: for ``What is the mobile phone number of the student named Timmothy Ward?'' the reference query filters on \texttt{first\_name = 'timmothy' and last\_name = 'ward'}, and the shipped database stores \texttt{Timmothy} and \texttt{Ward}, so the reference returns an empty table while the model's answer returns the number. A numeric comparison or ordering runs on a text-typed column: the \texttt{MPG} column of the car database is stored as text, so the reference query for ``Which model saves the most gasoline?'' orders lexicographically and returns \texttt{citroen}, while the model's answer casts to a number and returns \texttt{mazda}. A join without a join condition: the reference for ``What are the first name and last name of the professionals who have done treatment cheaper than average?'' joins professionals to treatments without an \texttt{ON} clause, returning fifteen names where the joined query returns five. A negated filter: the reference for ``What are the names of the dogs for which the owner spent more than 1000 for treatment?'' selects the dogs \texttt{NOT IN} the set that spent more than 1000. An incomplete projection: ``List the id and all lines'' of an address is answered with two of the three address-line columns. A set operation on names: ``students' first names who have both cats and dogs'' is answered by intersecting first names, which merges different students who share one. And a reversed or missing restriction: ``the weight of the youngest dog'' is answered by the youngest pet of any type. Some of these families disagree with the model's answer on the shipped database itself; others, such as the set operation on names and the missing pet-type restriction, return the same rows on the shipped database and are exposed only on suite instances that contain the collision. In every family the AI audit labelled the model's answer as the more faithful reading of the question.

\subsection{Mechanical signatures are associated with the label}
\label{sec:signatures}

The audit read the result tables of the reference query and the returned query on the first instance where they disagree, and a mechanical signature of that difference, computed before any label, is strongly associated with the label assigned in the disagreement census (\cref{fig:defects}b, appendix); this is an in-sample association with AI labels, not a validated predictor. When the two tables partially overlap, 93.8 percent of cases were labelled semantic errors (15 of 16); when the reference returns no rows on the instance, 53.3 percent; when the reference's rows are a subset of the returned rows, 34.4 percent; when the tables are disjoint, 25.0 percent (31 of 124). Five signatures never received the semantic-error label: differing duplicate multiplicity only (25 cases), the returned rows being a subset of the reference's (14), a different column count (9), the returned query returning no rows (3) and row order only (2), together 53 cases and 95 answers that the suite rejects and the audit attributes to the question, the reference or a synthetic instance rather than to the system. Disagreement breadth points the same way: cases that disagree on at most 10 percent of the instances were labelled semantic errors 17.9 percent of the time, cases that disagree on more than half of the instances 31.7 percent. Within the audited population, the signature lets a reviewer read the rejections most likely to be labelled semantic errors first.

\section{Re-scoring the certificate under the census labels}
\label{sec:reshapes}

 This section re-scores the certificate under the labels of \cref{sec:census}. The suite oracle stays the intervention throughout: cell $\cellD$ is still built and calibrated on the suite, and only the yardstick against which held-out answers are judged changes. Four nested conventions decide which rejected answers count as wrong: wide, the label is semantic error or synthetic-instance defect; narrow, the audit's final label is semantic error; agreed, both passes independently said semantic error; unanimous, agreed and neither pass called the case borderline. Everything the suite accepts is left as accepted, because no accepted answer was examined. All of it is post hoc and rests on those AI labels.

\subsection{The sign needs labels on both sides of the oracle}
\label{sec:sign}

\Cref{fig:hero}b and \cref{tab:gap_ladder} (appendix) give $\GAP$ under each convention. Relabelling the rejected side with the disagreement census alone leaves $\GAP$ positive, 2.26, 2.37, 2.54 and 1.00 points for Kwai-32B, Kwai-14B, Omni-32B and XiYan-32B under the narrow convention, down from 10.23, 8.08, 9.50 and 3.11; it cannot turn negative, because every convention keeps the suite label a refinement of the weak label on that population. Relabelling with the full census reverses the sign: $-4.99$, $-3.37$, $-4.25$ and $-4.22$ points under question splits and $-3.28$ to $-5.28$ under database splits. The mechanism is visible in the labels. The full census also covers answers that both oracles reject, and when the audit labels such an answer a suspected reference-query defect or an underspecified question, the shipped database has rejected it too; both oracles over-reject on those questions, and the shipped database's over-rejection inflates the reported risk the certificate is calibrated on, so that the certificate is conservative rather than optimistic under the audited yardstick. Under the narrow convention the current-practice cell of Kwai-32B carries 0.0482 at $\alpha = 0.10$ under question splits, three-seed mean, against 0.0981 under its own weak labels.

That reversal comes from auditing one side of the oracle. The census examined every answer the suite rejects and none that it accepts: cell $\cellA$ has 4,690 suite-accepted answer occurrences, 1,580 distinct question-and-SQL outputs, and a relabelling that only exculpates rejected answers can only lower the audited risk. Reference-query defects cut the other way as well: an answer that agrees with a defective reference on every instance is, if the label is right, evidence of being wrong, and 259 accepted occurrences (4.2 percent of all answers), 102 distinct outputs, fall on the 73 questions the audit flagged; counting all of them as wrong, an extreme bound given in \cref{tab:gap_ladder}, already restores a positive sign for XiYan-32B. 

\subsection{The D minus A contrast under audited labels}
\label{sec:ladder}

\Cref{tab:repair_ladder} gives $\DmA$ under the same conventions; its three-seed mean stays negative under every one. Under question splits the preregistered contrast of $-8.3$ to $-10.4$ points for three checkpoints shrinks to $-2.7$ to $-3.7$ under the narrow convention and to $-1.3$ to $-2.5$ under the unanimous one; for XiYan-32B it shrinks from $-2.9$ to $-1.0$ and $-0.7$. Under the wide convention all three clear the preregistered three-point bar under both split schemes. Among the narrow, agreed and unanimous conventions only Kwai-14B clears it under both split schemes, under narrow ($-3.74$ and $-3.65$) and agreed ($-3.36$ and $-3.27$); Kwai-32B and Omni-32B clear it under database splits only and only under narrow, and no checkpoint clears it under unanimous. The contrast is negative in 96.3 to 100 percent of splits for three checkpoints under every convention, on average over seeds; for XiYan-32B under database splits the share of negative splits falls from 0.702 under the preregistered yardstick to 0.437 under the unanimous one. The same ladder over the six scores of \cref{sec:sixscores} counts 32, 15, 10 and 1 of the 48 score-by-checkpoint-by-split cells above three points under the wide, narrow, agreed and unanimous conventions, so the shrinkage is not a property of top-class mass either. Answer rates are untouched by relabelling: the questions each cell answers are fixed by the calibration, and only their judged correctness changes.

\begin{table}[t]
\centering
\caption{The D minus A contrast (points, negative means cell D carries less risk under the named yardstick) at $\alpha=0.10$ under the preregistered suite labels and under four nested relabelling conventions of the full census that exculpate rejected answers labelled as not semantic errors: wide (semantic error and synthetic-instance defect count as wrong), narrow (only semantic error counts as wrong), agreed (narrow, restricted to cases both passes agreed on), unanimous (agreed and neither pass flagged the case as borderline). The suite oracle stays the intervention; only the yardstick changes. The right block is the three-seed mean share of the 200 splits in which the contrast is negative, a resplit-stability statistic. Over the six scores of Section~\ref{sec:sixscores}, the number of the 48 score-by-checkpoint-by-split cells that clear the preregistered three-point bar is 32, 15, 10 and 1 under the four conventions. Labels are AI-assigned. File-order recomputation (\cref{app:tie}).}
\label{tab:repair_ladder}
\footnotesize
\setlength{\tabcolsep}{3pt}
\begin{tabular}{llccccc ccccc}
\toprule
 & & \multicolumn{5}{c}{$\DmA$ (points)} & \multicolumn{5}{c}{share of splits negative} \\
\cmidrule(lr){3-7}\cmidrule(lr){8-12}
Checkpoint & Split & prereg. & wide & narrow & agreed & unanim. & prereg. & wide & narrow & agreed & unanim. \\
\midrule
Kwai-32B & question & -10.38 & -3.92 & -2.80 & -2.05 & -1.34 & 1.000 & 1.000 & 1.000 & 0.995 & 0.972 \\
 & database & -10.69 & -4.10 & -3.08 & -2.32 & -1.60 & 1.000 & 1.000 & 0.998 & 0.990 & 0.980 \\
Kwai-14B & question & -8.30 & -4.32 & -3.74 & -3.36 & -2.54 & 1.000 & 1.000 & 1.000 & 1.000 & 0.998 \\
 & database & -8.02 & -4.17 & -3.65 & -3.27 & -2.61 & 0.997 & 0.997 & 0.993 & 0.993 & 0.992 \\
Omni-32B & question & -9.58 & -3.85 & -2.71 & -2.25 & -1.35 & 1.000 & 1.000 & 1.000 & 0.998 & 0.963 \\
 & database & -10.21 & -4.13 & -3.14 & -2.68 & -1.68 & 1.000 & 0.997 & 0.988 & 0.982 & 0.963 \\
XiYan-32B & question & -2.90 & -1.66 & -1.05 & -0.99 & -0.68 & 0.973 & 0.943 & 0.803 & 0.800 & 0.710 \\
 & database & -1.25 & -0.73 & -0.43 & -0.46 & -0.35 & 0.702 & 0.577 & 0.485 & 0.473 & 0.437 \\
\bottomrule
\end{tabular}
\end{table}

\subsection{The contrast comes from which questions are answered}
\label{sec:triage}

The two partitions return the same SQL on 97.1 percent of answers (5,919 of 6,096): the intervention changes the returned query on 2.9 percent of question-and-pool pairs while it changes answer rates by 2.8 to 17.7 points. Decomposing $\DmA$ says where it comes from. Both cells are judged by the same oracle, so a held-out question on which both answer with the same SQL contributes exactly zero, and only three events can carry the contrast: both cells answer and the returned SQL differs, only cell $\cellD$ answers, or only cell $\cellA$ answers. Under question splits the changed-SQL term is $-0.001$, $0.000$, $-0.664$ and $0.005$ points for Kwai-32B, Kwai-14B, Omni-32B and XiYan-32B against an abstention term of $-10.379$, $-8.297$, $-8.911$ and $-2.907$, and under database splits the changed-SQL term reaches $-0.854$ for Omni-32B and stays below 0.03 for the other three (\cref{tab:dma_decomposition}). Changing which SQL is returned therefore carries 8.4 percent of the contrast at most, on Omni-32B under database splits, and under 0.2 percent for both Kwai checkpoints and for XiYan-32B under question splits; for XiYan-32B under database splits it offsets 2.1 percent of the contrast. The rest comes from questions only one cell answers, above all those cell $\cellA$ answers and cell $\cellD$ abstains on. Under the narrow convention, its contrast with cell $\cellA$ is $-1.05$ to $-3.74$ points under question splits, at answer rates 4.9 to 17.0 points lower; and under database splits XiYan-32B answers 2.8 points more questions in cell $\cellD$ than in cell $\cellA$ and its contrast is $-0.43$. The multi-instance cell therefore lowers risk by answering different questions rather than by returning different SQL, and its lower risk persists under expert labels (\cref{sec:expertrisk}).

\section{Neither oracle reports the risk an expert would assign}
\label{sec:experts}

Every certificate risk that \cref{sec:oracleswap,sec:reshapes} report is measured under one of the two oracles the benchmark ships or, in the audit ladder, under the suite oracle with its rejections re-judged by an AI reviewer. This section brings in a third reading. In a preregistered audit, two SQL experts judged whether a returned query answers its question, on every distinct answer of two checkpoints at generation seed 101, 1,028 items, blind to the cell, the score, the reference query and every oracle label, in a shuffled order, and required to execute rather than read. The two experts agree on the four-way verdict for 94.5 percent of the items and on correct against wrong for 97.9 percent, Cohen's $\kappa$ of 0.859 and 0.937. Their labels cover the answers the suite accepts as well as those it rejects, so they give the risk the certificate carries (\cref{sec:expertrisk}); they also test whether the usual evaluation of a consistency score can tell a better score from one aligned with the oracle that built it (\cref{sec:alignment,sec:yardstick}).

\subsection{The risk an expert assigns}
\label{sec:expertrisk}

The expert labels settle the question \cref{sec:sign} left open. This reading is post hoc: the preregistration fixed the alignment endpoint of \cref{sec:yardstick} and made no claim about the sign. The labels cover the answer each cell returns on every question, whether the suite accepted it or rejected it, so the sign of the semantic gap can be read off rather than bracketed. The experts call wrong 17.7 and 15.1 percent of the answers the suite accepts, on XiYan-32B and Kwai-32B, against 36.8 and 36.3 percent of the answers it rejects: the stricter oracle over-rejects, as the census of \cref{sec:census} found under AI labels, and it also accepts answers the experts do not. On the items both experts agree on, the shipped database agrees with them on 79.3 and 80.4 percent of answers, $\kappa$ of 0.199 and 0.306, and the suite oracle on 76.7 and 73.6 percent, $\kappa$ of 0.150 and 0.221: both oracles sit far from the experts, and the stricter one sits further on both checkpoints.

Re-scoring the certificate against these labels puts $\GAP$ at 10.10 and 7.37 points under the convention that falls back to the suite label where the experts left a question underspecified, and between 6.19 and 11.93 points over all nine combinations of the three fallback conventions with either expert's sheet or the agreed subset (\cref{tab:human_gap}). It is positive in every one of them. In levels rather than differences, the current-practice certificate at a nominal 0.10 carries 9.89 and 9.84 points of held-out risk under its own labels, 12.96 and 20.25 under the suite oracle, and 19.99 and 17.21 under the experts, on XiYan-32B and Kwai-32B. The shipped database understates the expert risk on both checkpoints, and the suite understates it by 7.03 points on one while overstating it by 3.04 on the other, so neither oracle is a conservative stand-in for the experts and the direction of the error is not a property of which oracle is stricter. On both checkpoints the current-practice certificate understates the risk an expert assigns it. The $\DmA$ contrast is negative in all eighteen cells as well, between $-1.42$ and $-4.57$ points, so the multi-instance cell carries less expert-judged risk too.

\subsection{Consistency scores are aligned with the oracle that built them}
\label{sec:alignment}

\begin{figure}[t]
    \centering
    \includegraphics[width=\textwidth]{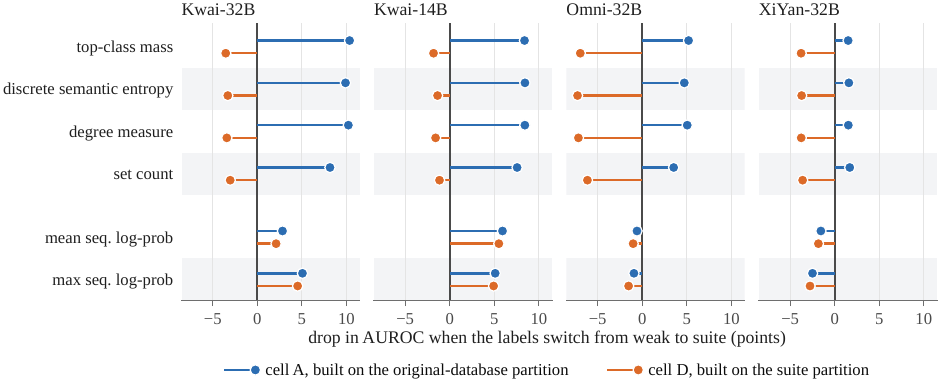}
    \caption{Oracle alignment. Each dot is the drop in AUROC from weak to suite labels, in points, for a score built on the original-database partition (cell $\cellA$, upper line) and for the same score built on the suite partition (cell $\cellD$, lower line). The four consistency scores drop under the other oracle's labels and gain under their own, so their two lines point in opposite directions in all 16 combinations; the two likelihood scores, computed by the same oracle-independent formula in both cells, never reverse. On the questions where both cells return the same SQL, all 16 reversals remain and each likelihood score is identical in both cells (\cref{tab:auroc}).}
    \label{fig:alignment}
\end{figure}

The oracle also bears on how an execution-consistency score is evaluated. Scores are usually assessed by their AUROC for predicting correctness, and \cref{tab:auroc} (appendix) gives the AUROC of the six scores under weak and suite labels on all 508 questions. Every score's AUROC changes when the labels change, and a drop under the stricter labels could mean only that the stricter labels are harder. A post-hoc symmetric comparison rules that reading out: build the same score once on the weak partition (cell $\cellA$) and once on the suite partition (cell $\cellD$), and switch the labels in both. A uniform label-difficulty shift would give the drop the same sign in both cells. It does not (\cref{fig:alignment}). For all four consistency scores and all four checkpoints, the weak-minus-suite AUROC change is positive when the score is built on the weak partition, 1.5 to 10.4 points, and negative when it is built on the suite partition, $-1.1$ to $-7.3$ points: each consistency score looks better under the labels of the oracle that built it, in 16 of 16 combinations, per seed as well as on average. The two likelihood scores, computed by the same oracle-independent formula in both cells, do not reverse. Restricting the computation to the questions on which both cells return the same SQL, 477 to 508 per seed, keeps all 16 reversals (1.70 to 9.82 points on cell $\cellA$, $-0.78$ to $-7.21$ on cell $\cellD$) and makes the likelihood scores exactly invariant, which is the null this test needs.

This is a signed oracle-by-partition interaction: it establishes descriptive oracle alignment in this setting and rules out a uniform label-difficulty shift. The observation is not a randomised intervention, and a substantive explanation remains open: the suite partition may capture the instability of a candidate set better and predict held-out suite correctness more accurately, while weak labels penalise it by counting its newly exposed errors as correct. The oracles' own labels do not separate that reading from a circular one. One consequence holds under either reading: an evaluation of an execution-consistency score under the labels of the oracle that produced its clusters cannot tell the two apart. On the strongest checkpoint the choice of labels changes the size of a lead: for XiYan-32B on the original-database partition the plain likelihood scores (0.704 and 0.702 under weak labels, 0.719 and 0.727 under suite labels) beat every consistency score (0.668 to 0.672 under weak labels, 0.653 to 0.656 under suite labels), and suite labels widen that lead from about three to about six points. The ordering is a property of the partition rather than of the scores: on the suite partition the same consistency scores reach 0.744 to 0.747 under suite labels, against 0.722 and 0.730 for the likelihood scores.

\subsection{An independent yardstick}
\label{sec:yardstick}

Both readings of \cref{sec:alignment} rest on evidence the two oracles produced, so the expert labels test whether the advantage they disagree about is visible outside them. The endpoint is the questions on which both cells return the same SQL: there one verdict serves both cells and the only thing that differs between them is the score attached to one answer. The decision rule was fixed before any label existed. The two checkpoints were chosen for opposite reasons, also fixed in advance: under the oracles' own labels the Kwai-32B score wins only under the oracle that built it, the symmetric pattern a circular explanation predicts, while the XiYan-32B score built on the suite partition wins under both sets of labels, the pattern it does not.

The rule returns mixed (\cref{tab:alignment_audit}). On the questions where the experts agree, the suite-partition score leads by 6.96 points of AUROC on Kwai-32B, $p = 0.004$, and by 1.53 points on XiYan-32B, $p = 0.275$. Each expert's own sheet returns the same pattern. The 57 items the experts disagree on are bracketed by a hill-climbing search rather than adjudicated: the resolutions that raise the contrast and those that lower it keep XiYan-32B between 0.47 and 2.72 points, never reaching significance, and Kwai-32B between 4.18 and 8.83 points, never losing it, and since the search is local these are the widest ranges it attained.

The comparison with the same contrast under suite labels is what the audit was for. Read on the same items the expert estimate is built from, the suite labels put the suite-partition score 8.3 points ahead on XiYan-32B and 9.2 ahead on Kwai-32B (\cref{app:yardstick}). Under expert labels the Kwai-32B lead holds, with an interval of 2.2 to 11.7 points that covers its suite-label value, and on XiYan-32B the expert interval, $-1.2$ to 4.3 points, excludes the 8.3 points the suite labels report. On the checkpoint chosen because its oracle-internal pattern looked least like alignment, most of the reported advantage is not visible under the independent labels; on the checkpoint chosen because it looked most like alignment, an advantage persists under a yardstick that did not build it. Neither outcome identifies a mechanism: a smaller real advantage that this audit cannot resolve is consistent with the first, and an oracle-specific component that still carries signal the experts agree with is consistent with the second. All four consistency scores give the same pattern, 1.5 to 1.7 points on XiYan-32B and 7.0 to 9.2 on Kwai-32B, so it is not a property of top-class mass, and the two likelihood scores, which read no oracle and take the same value in both cells, give exactly zero, which is the null this test needs.

The result is methodological. A lead the suite labels report can hold under an independent yardstick, as on Kwai-32B, or fall outside it, as on XiYan-32B; the two contrasts were not tested against each other. An improvement measured under the labels of the oracle that built a score's clusters is therefore not, on its own, evidence of an improvement that an independent reader would recognise.

\section{Discussion}
\label{sec:discussion}

The three results describe one dependence. The risk an execution-based certificate reports is a property of the oracle that labelled its calibration set, and in this setting neither oracle the benchmark publishes reports the risk an expert assigns. The measurement of \cref{sec:oracleswap} is large: it is the distance between the two oracles the benchmark publishes. \Cref{sec:census} shows what that distance is made of: most of it is attributed to the question, the reference query or a synthetic instance rather than to the system, and the reference-defect flag that carries the most weight is one the expert audit supports at the question level. Expert labels that cover both sides of the oracle fix the sign and the level of the certificate's risk (\cref{sec:expertrisk}). The same dependence reaches the evaluation of the scores. Each execution-consistency score looks better under the labels of the oracle that built its clusters (\cref{sec:alignment}), and under expert labels the lead the suite labels report holds on Kwai-32B and falls outside the expert interval on XiYan-32B (\cref{sec:yardstick}).

The leniency of the shipped database, on its own, would have predicted a simpler picture. The distilled test suite was introduced to correct that leniency \citep{zhong2020testsuite}, formal verification and synthesised databases have since confirmed it \citep{klopfenstein2025spotit,habibollah2026synsql}, and on that account the uncertainty pipelines that calibrate on the shipped database \citep{chen2026endtoend,maleki2025confidence} understate their risk by whatever the stricter oracle reveals. At the certificate layer the stricter oracle reveals errors and artefacts together, and the question of construct validity, whether a label produced by executing a reference query measures whether the question was answered, decides more of the reported number than leniency does. That is the observation \citet{klopfenstein2025spotit} made on fifty inspected counterexamples and \citet{pourreza2023evaluating} on the disagreements between models and benchmarks, carried to the layer that certifies a system's uncertainty and tested on the label that carries the most weight. The expert labels also place the certificate outside the regime in which label noise is benign: under dispersive noise a conformal certificate stays valid and errs on the conservative side \citep{einbinder2024labelnoise}, whereas on both audited checkpoints the certificate carries more expert-judged risk than it reports.

\subsection{Implications for practice}
\label{sec:practice}

A risk number without its oracle is not interpretable. The same certificate on the same pools carries a different held-out risk under each of the two oracles the benchmark publishes, the sign of the difference turns on whether the stricter oracle's rejections are believed, and where expert labels cover both sides the certificate that promises 0.10 carries 20.0 and 17.2 points, a figure neither oracle reports. In comparable text-to-SQL settings a certificate should be reported with the oracle that calibrated it and the oracle that evaluated it, and an oracle-relative difference should not be read as semantic risk until the benchmark's validity has been audited.

Multi-instance evaluation is not sufficient by itself in this setting, and it is not a conservative substitute either. It is the official metric and it is strictly stricter than the shipped database, yet the audit labelled most of its rejections as something other than a semantic error, and against expert labels it understates the certificate's risk on one checkpoint and overstates it on the other: being stricter does not fix the direction of the error. The audit an oracle-relative difference needs can start mechanically. The recurring families of \cref{sec:c7} are visible from executing the reference query alone: an empty result on the database it was written for, a comparison or ordering on a text-typed column, a join without a join condition; and in this population the difference signatures of \cref{sec:signatures} are associated with the assigned label strongly enough to prioritise which rejections a reviewer reads first.

For the evaluation of execution-consistency scores, \cref{sec:alignment} gives a concrete rule: where an independent oracle exists, evaluate a score under labels from an oracle that did not build its clusters and report both directions, because a single evaluation under the score's own oracle cannot distinguish a better score from an aligned one.

\subsection{Limitations}
\label{sec:limitations}

\paragraph{What the evidence is relative to.} The 73 flagged questions are the ones four particular checkpoints exposed rather than a sample. The sign of the semantic gap is identified where both sides of the oracle were labelled, which is two of the four checkpoints at one generation seed; on the other two, and on the other seeds, the accepted side is unexamined. On official Spider dev, where most questions place every candidate in one class, the preregistered control gave XiYan-32B a $\DmA$ contrast of $-0.64$ points, below the three-point criterion (\cref{app:history}).

\paragraph{Labels.} Human SQL experts have tested one of the five census labels, the suspected reference-query defect, and at the question level (\cref{app:humanaudit}): the case-level attributions of the census, which pair a reference query with a particular returned answer, were never shown to the experts, and the semantic-error, underspecified-question, synthetic-instance and comparator-artefact labels remain AI-only. Human verdicts and AI flags agree on 63.7 percent of the 113 audited questions the census had examined, $\kappa = 0.305$, and on 71.2 percent of all 153, $\kappa = 0.416$, when the 40 never-examined ones count as AI negatives; both agreement rates are below the experts' 94.1 percent binary agreement on those 153, so the flag discriminates without being the same judgement. The two experts in the reference-query audit share a laboratory with the authors, the project lead adjudicated their thirteen four-way disagreements, and the protocol left them aware of the hypothesis and illustrated defect families the AI audit had produced. The preregistered rule returns supported on either expert's sheet alone, at margins of 36.2 and 27.7 points against 34.8 after adjudication; the illustrations' influence on their judgements is unmeasured. The label taxonomy was defined after the disagreement census was inspected, labellers saw the result tables on the first disagreeing instance only, and the prompts and call traces of the labelling passes were not retained, so the evidence that the two passes were independent is their recorded metadata. The expert audit of \cref{sec:experts} was run by two other experts under a protocol that illustrates no defect family, and its labels are their judgements of whether a returned answer answers its question (\cref{app:yardstick}).

\paragraph{Generality.} The study covers one benchmark family on SQLite with one distilled suite, and two Qwen lineages of one vendor, because no non-Qwen SQL specialist available to us passed the frozen entry rule; other vendors and benchmark families are untested, and the matched original-Spider control covers one lineage. 

\paragraph{Registration.} The third registration names XiYan-32B as the Qwen2.5-Coder representative while the order of its candidate table would name Omni-32B, and the paper adopts the conservative reading, under which the $\DmA$ contrast is not cross-lineage (\cref{sec:prereg,app:history}). The endpoint and the benchmark, Spider-Realistic, were chosen during exploratory development with XiYan-32B, after an original selective-risk endpoint on official Spider dev missed its three-point criterion; both were fixed in the first registration, and the third registration froze the four-checkpoint panel analysis before any of its main or control pools were generated (\cref{app:history}). 

\subsection{Future work}
\label{sec:future}

Auditing the accepted side on the rest of the panel, by the protocol of \cref{app:yardstick}, would extend the expert-judged level to every checkpoint and seed. Comparing cells $\cellA$ and $\cellD$ at equal answer rate under expert labels would measure what multi-instance evaluation buys once abstention is held fixed. Replication across vendors, and across benchmark families once another benchmark ships a multi-instance oracle of comparable quality, would tell whether the families of \cref{sec:c7} are Spider's or the field's.

\Needspace*{6\baselineskip}
\section{Conclusion}
\label{sec:conclusion}

The oracle that labels the calibration set decides the risk an execution-based conformal abstention certificate for text-to-SQL reports. On Spider-Realistic, the current-practice certificate carries 2.73 to 10.23 points more held-out risk under the multi-instance suite oracle than under its shipped-database labels, on four SQL-specialist checkpoints. Most of that difference lies in the benchmark: an AI-assigned census attributes most of what the stricter oracle rejects to the benchmark's questions, reference queries and synthetic instances rather than to the system, and two SQL experts supported the reference-defect flag in a preregistered blinded audit. Expert labels that cover both sides of the oracle give the expert-judged risk: the certificate that promises 0.10 carries 20.0 and 17.2 points on the two checkpoints audited, which neither oracle reports. The shipped database understates it on both, and the suite understates it on one and overstates it on the other: being stricter does not fix the direction of the error. The same dependence governs how a score is judged: an execution-consistency score looks better under the labels of the oracle that built its clusters, in every combination tested, and on one of the two audited checkpoints the lead the suite labels report falls outside the expert interval.

A certificate is therefore a statement about an oracle and should be reported as one, with the oracle that calibrated it and the oracle that evaluated it. An oracle-relative difference should be read as semantic risk only after the benchmark's validity has been audited, and a consistency score should be evaluated under an oracle that did not build it.

\bibliographystyle{plainnat}
\bibliography{references}

\clearpage
\appendix
\setcounter{figure}{0}
\setcounter{table}{0}
\renewcommand{\thefigure}{A\arabic{figure}}
\renewcommand{\thetable}{A\arabic{table}}
\phantomsection\addcontentsline{toc}{section}{Appendix}%
\noindent{\Large\bfseries Appendix}\par
\section{Endpoint evolution and preregistration history}
\label{app:history}

Three preregistrations preceded the final preregistered panel. \Cref{tab:prereg} lists what each froze and what it found. Apart from the analyses that the two expert audits preregistered (\cref{app:humanaudit,app:yardstick}), every analysis after the preregistered result of \cref{tab:main} is post hoc. Each preregistration document carries the content hashes of the frozen code and data together with its deviation record. Each registration was written after the results of the previous stage were known and before any of its own pools was generated; the twelve main pools and four control pools of the panel were generated after the third registration and had not been sampled before it. The endpoint and the benchmark were therefore chosen during exploratory development on the same benchmark with one checkpoint, and the final panel is a fresh-generation confirmation of that endpoint on four checkpoints. The registrations name the D minus A contrast REPAIR.

\begin{table}[htbp]
\centering
\caption{The registration history of the study. Each row froze its data, models, endpoint and decision rule before its pools were generated; the last column reports what the frozen rule returned.}
\label{tab:prereg}
\footnotesize
\setlength{\tabcolsep}{4pt}
\begin{tabular}{>{\raggedright\arraybackslash}p{0.15\linewidth} >{\raggedright\arraybackslash}p{0.29\linewidth} >{\raggedright\arraybackslash}p{0.23\linewidth} >{\raggedright\arraybackslash}p{0.23\linewidth}}
\toprule
Stage & Data and models & Endpoint and rule & Result \\
\midrule
Original proposal (before any registration) & Official Spider dev, 1,034 questions, XiYanSQL-QwenCoder-32B-2504 & Multi-instance intervention lowers selective risk by at least three points & Reduction 2.42 to 2.75 points; criterion not met; endpoint replaced by the marginal-risk quantities $\GAP$ and $\DmA$ on Spider-Realistic \\
Registration 1 (2026-09-04) & Spider-Realistic, 508 questions, the same checkpoint, official comparator, three new generation seeds (four pools with the existing one) & $\GAP > 0$ in every pool; $\DmA \le -3$ points & $\GAP$ 3.19 to 3.52 points in all four pools; $\DmA$ crossed the bar in two of four pools \\
Registration 2 (2026-09-04) & Five general-purpose candidates under an entry rule; database-grouped splits; false-schema control & Entry rule, then the same endpoints on every entrant & No general candidate entered (outcome O5); on the XiYanSQL checkpoint grouped splits kept $\GAP$ and halved $\DmA$; the control passed \\
Registration 3 (2026-09-05) & Six SQL-specialist candidates, disjoint 200-question entry pilot, seeds 101, 202 and 303, question and database splits co-primary, aggregation rule and outcome table (\cref{sec:prereg_rules}) & $\GAP$ rule and $\DmA$ rule of \cref{sec:prereg_rules} & Four entrants; outcome O2 (\cref{tab:main}) \\
Post hoc & Same pools & Six-score extension, robustness checks, both censuses, both ladders, the D minus A decomposition and the alignment finding & \cref{sec:oracleswap,sec:census,sec:reshapes} \\
\bottomrule
\end{tabular}
\end{table}

\paragraph{The saturated regime.} On official Spider dev with XiYanSQL-QwenCoder-32B-2504, the score saturates: in a 500-question pilot with 20 samples per question, 465 questions placed every sample in one execution class and the score took 11 distinct values. The preregistered control on all 1,034 dev questions with 50 samples changed suite-oracle marginal risk by $-0.64$ points (D minus A) and coverage by $+2.17$ points, with 5.1 percent of questions having a shipped-database class that the suite splits; both changes are below the three-point criterion, and the control tested whether the sample budget produces the Spider-Realistic effect: raising the budget from 20 to 50 samples moved the dev-set contrast from 0.26 to 0.64 points while the Spider-Realistic contrast on the same checkpoint was unchanged.

\paragraph{Deviations from Registration 3.} Two deviations are on record: the candidate table listed OmniSQL-32B before XiYanSQL-QwenCoder-32B-2504 while the text named the latter as the lineage representative, so a literal reading of the table order gives outcome O1, and both readings are reported with O2 as the conservative one; and the OmniSQL-32B pools were generated with tensor parallelism two where the frozen configuration said four, a serving choice.

\FloatBarrier
\section{The tie-rule audit}
\label{app:tie}

\begin{table}[htbp]
\centering
\caption{The tie-rule audit. The frozen analyser breaks a top-class tie by the largest integer union-find key; the per-question files sort by the stringified key, so every recomputation from them takes the first-listed class. Impact bounds are over every reported GAP and D minus A cell after recomputing under the frozen rule and bracketing the never-labelled answers in both directions.}
\label{tab:tie}
\small
\begin{tabular}{lc}
\toprule
Top-class decisions examined & 12,192 \\
Tied decisions & 53 \\
Tied decisions whose representative differs under the frozen rule & 51 \\
Tied decisions whose correctness label differs & 27 \\
Disagreement census: archived / frozen answers & 521 / 522 \\
Full census: archived / frozen answers (cases) & 2,779 / 2,771 (742 / 740) \\
Largest change of any reported GAP or D minus A (points) & 0.0841 \\
Largest bracket width from never-labelled answers (points) & 0.0631 \\
Sign changes / crossings of the three-point line & 0 / 0 \\
\bottomrule
\end{tabular}
\end{table}

The frozen analyser and the archived per-question files break a tie for the largest execution class differently (\cref{sec:designchoices}), and every recomputation from the files, including both censuses and all audited ladders, uses the file-order rule. A released script reconstructs the frozen analyser's own choice on each tied decision from the per-question files and applies it, so a recomputation can take either convention rather than being confined to the one the file order forces. \Cref{tab:tie} quantifies the consequence. When the frozen choice is reconstructed by re-running the union-find on the tied questions only, 53 of 12,192 top-class decisions are tied; 51 of them return a different SQL under the frozen rule and 27 change a correctness label. The frozen populations are therefore 522 answers for the disagreement census, one case the audit never saw and one labelled case outside the population, and 2,771 answers over 740 cases for the full census, 15 cases never seen and 17 labelled cases outside. Recomputing every reported $\GAP$ and $\DmA$ under the frozen rule, with the never-labelled answers bracketed as all semantic errors and as none, changes no cell by more than 0.0841 points; the two brackets differ by at most 0.0631 points; no sign changes and no cell crosses the three-point line, although 83 of the 9,600 split-level thresholds and 66 answer rates do move. The preregistered numbers of \cref{tab:main} are the frozen analyser's; the audited ladders of \cref{tab:repair_ladder,tab:gap_ladder} are file-order recomputations, which is why their preregistered columns differ from it. The 0.0841 bound is on the unrounded means, and the two-decimal values the tables print differ by up to 0.09 points.

\FloatBarrier
\section{Census details}
\label{app:census}

\begin{table}[htbp]
\centering
\caption{Labels assigned to the two censuses, by distinct case and by answer occurrence. Labels are AI-assigned. The lower block gives the error type of the cases labelled semantic error.}
\label{tab:labels}
\small
\begin{tabular}{lcccc}
\toprule
 & \multicolumn{2}{c}{Disagreement census} & \multicolumn{2}{c}{Full census} \\
\cmidrule(lr){2-3}\cmidrule(lr){4-5}
Label & cases (240) & answers (521) & cases (742) & answers (2,779) \\
\midrule
semantic error & 65 (27.1\%) & 136 (26.1\%) & 288 (38.8\%) & 972 (35.0\%) \\
suspected reference-query defect & 42 (17.5\%) & 107 (20.5\%) & 213 (28.7\%) & 887 (31.9\%) \\
underspecified question & 93 (38.8\%) & 195 (37.4\%) & 184 (24.8\%) & 723 (26.0\%) \\
synthetic-instance defect & 40 (16.7\%) & 83 (15.9\%) & 47 (6.3\%) & 161 (5.8\%) \\
comparator artefact & 0 (0.0\%) & 0 (0.0\%) & 10 (1.3\%) & 36 (1.3\%) \\
\midrule
Error type of semantic-error cases & & & & \\
\quad wrong filter & 24 & & 129 & \\
\quad wrong grouping & 24 & & 30 & \\
\quad wrong join & 3 & & 44 & \\
\quad wrong aggregate & 4 & & 31 & \\
\quad wrong set operation & 4 & & 26 & \\
\quad wrong order or limit & 0 & & 19 & \\
\quad wrong projection & 6 & & 9 & \\
\bottomrule
\end{tabular}
\end{table}

\begin{figure}[htbp]
    \centering
    \includegraphics[width=\textwidth]{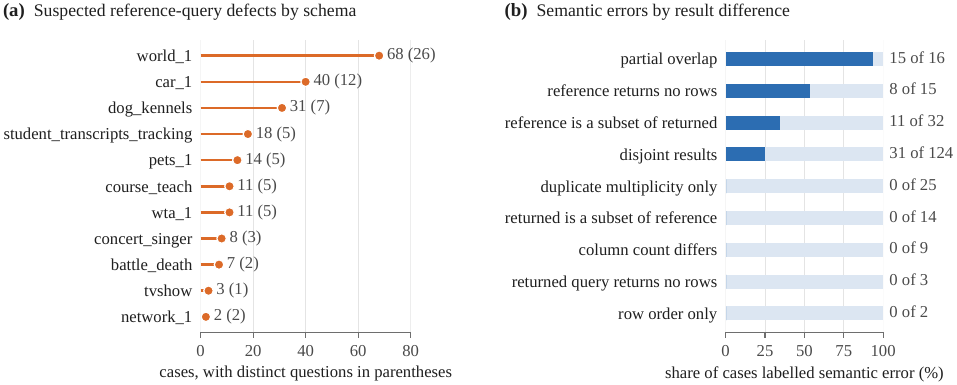}
    \caption{(a) Cases the AI audit labelled as a suspected reference-query defect in the full census, by schema, with the number of distinct questions in parentheses; the human audit of \cref{tab:blind} gives a post-hoc design-weighted estimate of 17.5 percent of the 508 questions. (b) Share of cases labelled semantic error by the mechanical signature of the result difference on the first disagreeing instance, disagreement census; the signature is computed before any label is assigned.}
    \label{fig:defects}
\end{figure}

\paragraph{Taxonomy.} Semantic error: the returned query answers the natural-language question wrongly. Underspecified question: the question does not fix the point on which the two queries differ, and both readings are faithful to it; typical cases are ties at an extremum where \texttt{ORDER BY ... LIMIT 1} and \texttt{WHERE x = (SELECT MIN(x) ...)} differ, disagreement over whether to deduplicate, and ambiguity over whether ``name'' means a first name or a full name. Suspected reference-query defect: the reference query is, in the labeller's judgement, the worse rendering of the question. Synthetic-instance defect: the two queries differ only on values a real database of the schema would not contain, such as one country name mapped to several codes. Comparator artefact: the two results carry the same information and the comparator still judges them unequal. For semantic errors the labeller recorded one of wrong filter, wrong join, wrong aggregate, wrong grouping, wrong set operation, wrong order or limit, and wrong projection.

\paragraph{Batching and adjudication.} The disagreement census was labelled in six batches of forty consecutive cases in the first pass and re-batched with a stride of six, with the taxonomy listed in reverse order, in the second; the full census used a stride of thirteen. The second pass on the full census was additionally told that 472 of the 742 cases already disagree on the shipped database, so that a synthetic-instance-defect label there would amount to calling the benchmark's own database defective, a stronger claim to be made only with reason. A third fresh thread saw both labels and both rationales for every disagreement and could rule for either side or against both. Each pass could flag a case as borderline; 165 of 240 disagreement-census cases and 332 of 742 full-census cases carry at least one flag.

\paragraph{Distribution by checkpoint and schema.} In the full census the semantic-error share of each checkpoint's audited answers is 30.1 percent for Kwai-32B, 41.5 for Kwai-14B, 29.4 for Omni-32B and 42.8 for XiYan-32B. Suspected reference-defect cases concentrate in world\_1 (68 cases, 26 questions), car\_1 (40, 12) and dog\_kennels (31, 7), with 11 schemas represented; \cref{fig:defects}a gives the full distribution. Of the 288 semantic-error cases, 129 are wrong filters, 44 wrong joins, 31 wrong aggregates, 30 wrong groupings, 26 wrong set operations, 19 wrong orderings or limits and 9 wrong projections.

\FloatBarrier
\section{Instrument checks}
\label{app:instrument}

\begin{figure}[htbp]
    \centering
    \includegraphics[width=\textwidth]{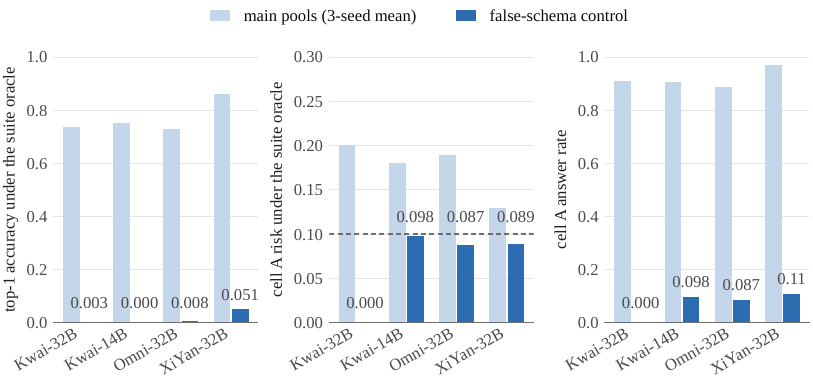}
    \caption{False-schema negative control. Each checkpoint was prompted with the schema of an unrelated database, one control pool per checkpoint with 20 samples per question and generation seed 999, against the three-seed means of the main pools with 50 samples per question; the control pools have 303, 481, 508 and 508 usable questions for Kwai-32B, Kwai-14B, Omni-32B and XiYan-32B, so the comparison is a diagnostic rather than a matched effect estimate. Top-1 accuracy under the suite oracle, the suite-oracle risk of cell A and its answer rate are shown. The certificate abstains on 89 to 100 percent of questions and keeps the suite-oracle risk at or below the nominal level plus 0.02, the preregistered expectation.}
    \label{fig:negctl}
\end{figure}

Every main pool has a parse rate of at least 0.9985 and no truncated generation (\cref{tab:panel}). Under the false schema the suite top-1 accuracy of Kwai-32B, Kwai-14B, Omni-32B and XiYan-32B is 0.0033, 0.0, 0.0079 and 0.0512; the two Kwai checkpoints mostly decline to write SQL (parse rates 0.156 and 0.325); Omni-32B still writes a query for 70.7 percent of samples and XiYan-32B for 99.98 percent; and the current-practice certificate answers 0.0, 9.8, 8.7 and 11.0 percent of questions with a suite-oracle marginal risk of 0.0, 0.098, 0.087 and 0.089, within the preregistered expectation of $\alpha + 0.02$. Two sensitivity analyses accompany the preregistered result: under the budget denominator the question-split $\GAP$ and $\DmA$ change by at most 0.14 points while some secondary operating points change substantially, because unusable samples become implicit abstention mass; and under a custom canonicalising comparator in place of the official one the direction is unchanged and the magnitudes slightly smaller, for example $-10.1$ to $-9.7$ points of contrast for Kwai-32B against $-10.4$. Refining the threshold grids of all six scores five-fold changes $\GAP$ and $\DmA$ by at most 0.13 points.

The gold-free construction reads the row-order flag per compared pair, which makes the equivalence relation non-transitive on at most 11 questions per pool; the union-find takes the transitive closure, so a gold-free class is a connected component rather than a set of pairwise equivalent candidates. The candidates-only construction reads one flag per question and is therefore transitive, and it differs from the gold-free construction in nothing else, so the gap between the two, at most 0.01 points on any checkpoint, is the net sensitivity to reading the flag per pair together with the closure that reading forces.

\section{Additional figures and tables}
\label{app:more}

\begin{figure}[htbp]
    \centering
    \includegraphics[width=\textwidth]{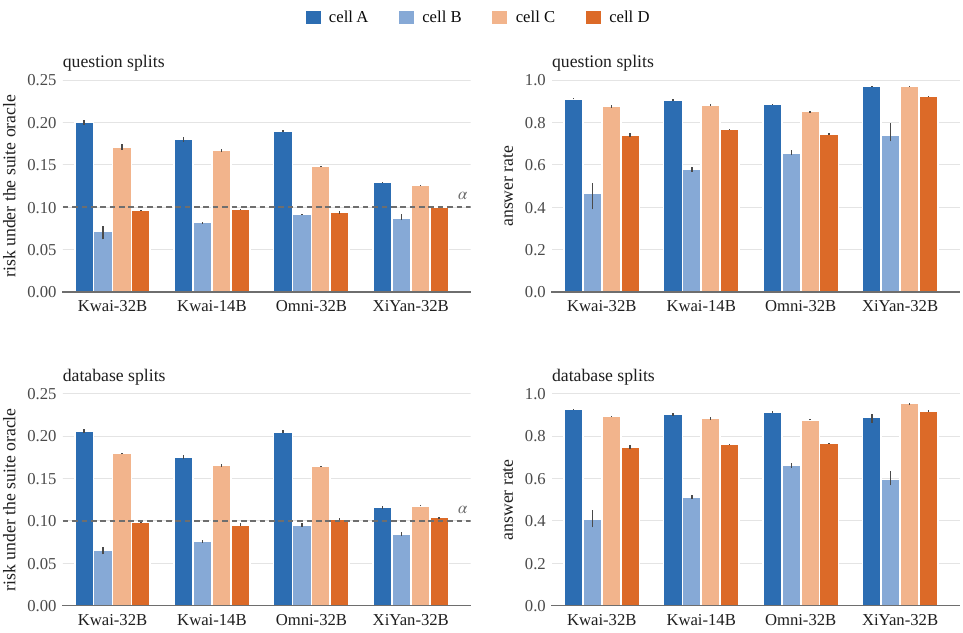}
    \caption{Risk under the multi-instance suite oracle and answer rate of the four cells of the intervention at nominal risk 0.10, three-seed means with the range over seeds, under question splits (top) and database splits (bottom). Cell A is current practice (original-database partition and labels) and cell D is fully multi-instance; cells B and C change only the labels or only the partition. The dashed line marks the nominal level.}
    \label{fig:cells}
\end{figure}

\begin{figure}[htbp]
    \centering
    \includegraphics[width=\textwidth]{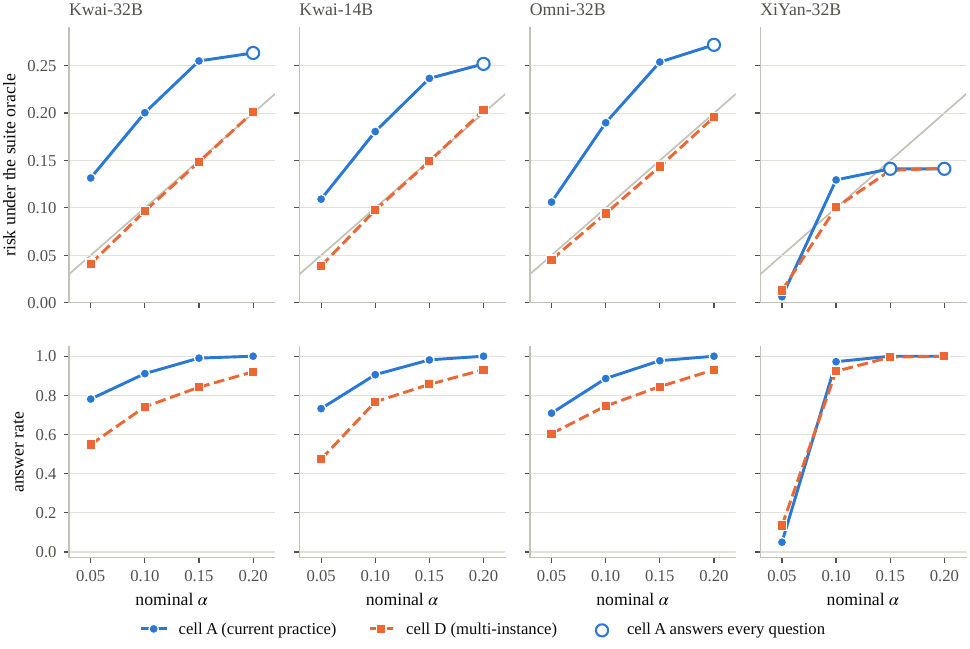}
    \caption{Suite-oracle risk (top) and answer rate (bottom) of the current-practice cell A and the multi-instance cell D against the nominal level, question splits, three-seed means. The grey line is the identity. Hollow markers show levels at which cell A answers every question, where the certificate has degenerated to answering everything and the ratio to the nominal level necessarily shrinks.}
    \label{fig:alpha}
\end{figure}

\begin{figure}[htbp]
    \centering
    \includegraphics[width=\textwidth]{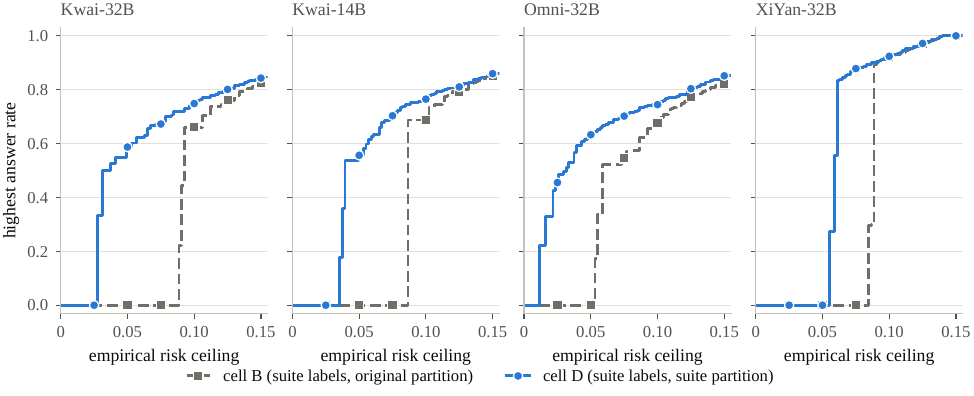}
    \caption{Highest answer rate reachable under a common empirical risk ceiling by cell B (suite labels, original-database partition) and cell D (suite labels, suite partition), three-seed means over all thresholds and the abstain-on-everything point. This is an empirical frontier on the full data, the two cells satisfy the same ceiling rather than the same realised risk, and it is not a held-out guarantee. The lines trace the frontier at every ceiling, and the markers give its value at the ceilings 0.025 to 0.15 in steps of 0.025.}
    \label{fig:frontier}
\end{figure}

\begin{figure}[htbp]
    \centering
    \includegraphics[width=\textwidth]{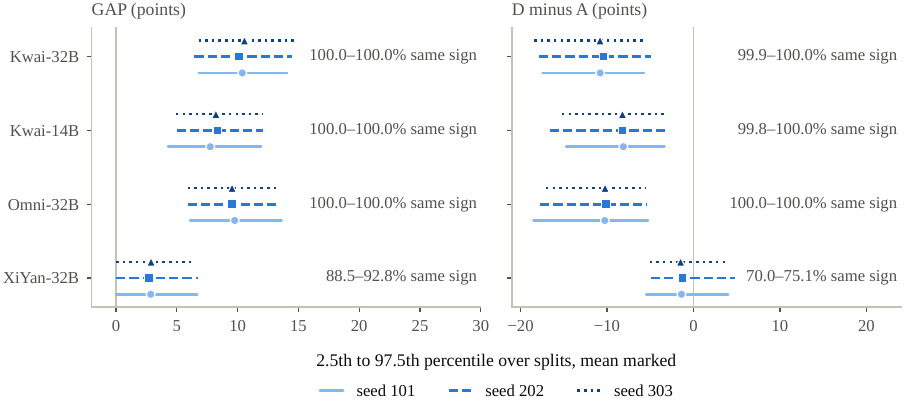}
    \caption{GAP and D minus A over all 92,378 schema-disjoint 9/10 splits of the 19 schemas, per checkpoint and generation seed: mean and the 2.5th to 97.5th percentile range over splits, with the range over seeds of the share of splits carrying the reported sign.}
    \label{fig:exhaustive}
\end{figure}

This section collects the four cells of the intervention (\cref{fig:cells}), the sweep over nominal levels (\cref{fig:alpha}), the empirical frontier (\cref{fig:frontier}), the exhaustive schema splits (\cref{fig:exhaustive}) and the appendix tables referenced from the main text.

\begin{table}[htbp]
\centering
\caption{GAP (points) at $\alpha=0.10$: the risk of the current-practice cell under the named yardstick minus the risk it reports on its own labels; only the preregistered column uses the unmodified suite oracle as the yardstick. Columns: under the preregistered suite labels, after relabelling the rejected side with the disagreement census (narrow convention), after relabelling it with the full census (narrow and unanimous conventions), and under the reverse stress test that additionally counts as wrong the 259 accepted answers that fall on questions with a suspected defective reference query. The stress test is an extreme assumption, not an estimate; the 1,580 distinct accepted outputs were never examined. Labels are AI-assigned. File-order recomputation (\cref{app:tie}).}
\label{tab:gap_ladder}
\footnotesize
\setlength{\tabcolsep}{4pt}
\begin{tabular}{llccccc}
\toprule
Checkpoint & Split & preregistered & disagreement census & full census, narrow & full census, unanimous & reverse stress \\
\midrule
Kwai-32B & question & +10.23 & +2.26 & -4.99 & -7.51 & -3.22 \\
 & database & +10.17 & +2.21 & -5.28 & -7.81 & -3.38 \\
Kwai-14B & question & +8.08 & +2.37 & -3.37 & -5.90 & -0.24 \\
 & database & +7.87 & +2.27 & -3.28 & -5.59 & -0.26 \\
Omni-32B & question & +9.50 & +2.54 & -4.25 & -6.55 & -2.18 \\
 & database & +9.49 & +2.55 & -5.17 & -7.63 & -2.91 \\
XiYan-32B & question & +3.11 & +1.00 & -4.22 & -7.15 & +3.20 \\
 & database & +2.73 & +0.81 & -4.08 & -6.51 & +2.72 \\
\bottomrule
\end{tabular}
\end{table}

\begin{table}[htbp]
\centering
\caption{GAP and D minus A (points) for six confidence scores under one certificate, one answer rule, the same pools and the same 200 splits, at $\alpha=0.10$ under the preregistered suite labels. The top-class-mass row is the pipeline of the main text; the four consistency scores share the execution-equivalence relation under test; the two likelihood scores never read any oracle. All 48 GAP cells are positive and all 48 D minus A cells are negative. File-order recomputation (\cref{app:tie}).}
\label{tab:sixscores}
\small
\setlength{\tabcolsep}{2.5pt}
\begin{tabular}{l cccc cccc}
\toprule
 & \multicolumn{4}{c}{question splits} & \multicolumn{4}{c}{database splits} \\
\cmidrule(lr){2-5}\cmidrule(lr){6-9}
Score & Kwai-32B & Kwai-14B & Omni-32B & XiYan-32B & Kwai-32B & Kwai-14B & Omni-32B & XiYan-32B \\
\midrule
\multicolumn{9}{l}{\emph{GAP (points)}} \\
top-class mass & +10.2 & +8.1 & +9.5 & +3.1 & +10.2 & +7.9 & +9.5 & +2.7 \\
discrete semantic entropy & +9.8 & +8.1 & +8.7 & +3.2 & +9.7 & +7.9 & +9.0 & +2.8 \\
degree measure & +10.1 & +8.1 & +9.2 & +3.1 & +10.1 & +7.9 & +9.4 & +2.8 \\
set count & +8.8 & +7.4 & +7.7 & +3.2 & +8.7 & +7.1 & +8.1 & +2.8 \\
mean seq. log-prob & +8.4 & +7.1 & +6.7 & +3.1 & +8.3 & +6.8 & +7.4 & +2.8 \\
max seq. log-prob & +9.1 & +6.6 & +6.8 & +3.0 & +9.1 & +6.4 & +7.5 & +2.7 \\
\midrule
\multicolumn{9}{l}{\emph{$\DmA$ (points)}} \\
top-class mass & -10.4 & -8.3 & -9.6 & -2.9 & -10.7 & -8.0 & -10.2 & -1.2 \\
discrete semantic entropy & -10.0 & -8.2 & -9.0 & -2.9 & -10.4 & -8.0 & -9.6 & -1.2 \\
degree measure & -10.2 & -8.2 & -9.3 & -2.9 & -10.5 & -7.9 & -10.0 & -1.2 \\
set count & -11.8 & -11.5 & -8.5 & -3.7 & -11.7 & -10.3 & -9.5 & -2.0 \\
mean seq. log-prob & -8.5 & -7.3 & -7.0 & -2.9 & -8.9 & -7.1 & -7.8 & -1.6 \\
max seq. log-prob & -9.3 & -6.8 & -7.1 & -3.0 & -9.9 & -6.9 & -7.8 & -1.4 \\
\bottomrule
\end{tabular}
\end{table}

\begin{table}[htbp]
\centering
\caption{AUROC of each score for predicting the correctness of the returned answer, computed on all 508 questions of each seed and averaged over the three seeds; the value does not depend on the split scheme. The score is built once on the original-database partition (cell A) and once on the suite partition (cell D), and each is evaluated under weak labels (correctness on the shipped database) and under suite labels. The drop is the weak-minus-suite change in AUROC in points for each cell; the same-SQL drop restricts the computation to the questions on which both cells return the same SQL, 477 to 508 per seed. The two likelihood scores use the same oracle-independent formula in both cells.}
\label{tab:auroc}
\footnotesize
\setlength{\tabcolsep}{4.5pt}
\begin{tabular}{ll cc cc cc cc}
\toprule
 & & \multicolumn{2}{c}{Built on A} & \multicolumn{2}{c}{Built on D} & \multicolumn{2}{c}{Drop} & \multicolumn{2}{c}{Same-SQL drop} \\
\cmidrule(lr){3-4}\cmidrule(lr){5-6}\cmidrule(lr){7-8}\cmidrule(lr){9-10}
Checkpoint & Score & weak & suite & weak & suite & A & D & A & D \\
\midrule
Kwai-32B & top-class mass & 0.846 & 0.743 & 0.804 & 0.839 & +10.4 & -3.5 & +9.8 & -3.1 \\
 & discrete semantic entropy & 0.842 & 0.743 & 0.804 & 0.837 & +9.9 & -3.3 & +9.4 & -3.0 \\
 & degree measure & 0.845 & 0.743 & 0.805 & 0.840 & +10.2 & -3.4 & +9.7 & -3.1 \\
 & set count & 0.815 & 0.733 & 0.777 & 0.807 & +8.2 & -3.0 & +7.8 & -2.9 \\
 & mean seq. log-prob & 0.711 & 0.682 & 0.710 & 0.689 & +2.8 & +2.1 & +2.0 & +2.0 \\
 & max seq. log-prob & 0.712 & 0.661 & 0.712 & 0.666 & +5.1 & +4.5 & +4.4 & +4.4 \\
\addlinespace
Kwai-14B & top-class mass & 0.838 & 0.754 & 0.810 & 0.828 & +8.4 & -1.8 & +8.7 & -1.0 \\
 & discrete semantic entropy & 0.839 & 0.754 & 0.812 & 0.826 & +8.4 & -1.4 & +8.8 & -0.8 \\
 & degree measure & 0.838 & 0.754 & 0.813 & 0.829 & +8.4 & -1.6 & +8.8 & -0.9 \\
 & set count & 0.819 & 0.744 & 0.787 & 0.799 & +7.6 & -1.1 & +7.8 & -1.1 \\
 & mean seq. log-prob & 0.740 & 0.681 & 0.727 & 0.672 & +5.9 & +5.5 & +6.0 & +6.0 \\
 & max seq. log-prob & 0.696 & 0.645 & 0.684 & 0.634 & +5.1 & +4.9 & +5.0 & +5.0 \\
\addlinespace
Omni-32B & top-class mass & 0.834 & 0.782 & 0.780 & 0.850 & +5.2 & -7.0 & +4.2 & -6.9 \\
 & discrete semantic entropy & 0.831 & 0.783 & 0.777 & 0.849 & +4.7 & -7.3 & +3.7 & -7.2 \\
 & degree measure & 0.833 & 0.783 & 0.779 & 0.851 & +5.1 & -7.2 & +4.0 & -7.1 \\
 & set count & 0.802 & 0.767 & 0.753 & 0.814 & +3.5 & -6.2 & +2.4 & -6.0 \\
 & mean seq. log-prob & 0.651 & 0.657 & 0.644 & 0.655 & -0.6 & -1.0 & -1.1 & -1.1 \\
 & max seq. log-prob & 0.667 & 0.677 & 0.657 & 0.672 & -0.9 & -1.5 & -1.4 & -1.4 \\
\addlinespace
XiYan-32B & top-class mass & 0.668 & 0.653 & 0.706 & 0.744 & +1.5 & -3.8 & +1.7 & -3.5 \\
 & discrete semantic entropy & 0.670 & 0.655 & 0.710 & 0.747 & +1.6 & -3.7 & +1.8 & -3.4 \\
 & degree measure & 0.669 & 0.653 & 0.707 & 0.745 & +1.5 & -3.8 & +1.7 & -3.5 \\
 & set count & 0.672 & 0.656 & 0.711 & 0.747 & +1.7 & -3.6 & +1.9 & -3.4 \\
 & mean seq. log-prob & 0.704 & 0.719 & 0.704 & 0.722 & -1.6 & -1.8 & -1.5 & -1.5 \\
 & max seq. log-prob & 0.702 & 0.727 & 0.702 & 0.730 & -2.5 & -2.8 & -2.4 & -2.4 \\
\bottomrule
\end{tabular}
\end{table}

\begin{table}[htbp]
\centering
\caption{Post-hoc robustness checks of the preregistered measurement (points, three-seed means, $\alpha=0.10$, question splits unless stated). Instance cross-fit: classes and calibration labels from one half of each schema's suite instances plus the shipped database, correctness judged on the other half plus the shipped database, both directions (24 runs); the held-out D-cell risk row is its range. Candidates-only: partition built from the sampled candidates alone, the reference query used only to label a class afterwards. Gold-free: candidates-only with the row-order convention taken from the compared pair. Exhaustive database splits: mean over all 92,378 schema-disjoint 9/10 splits, with the range over seeds of the share of splits carrying the reported sign given in the two rows below. Leave-one-database-out: the preregistered question-split analysis with one schema removed. The matched original-Spider control (Kwai-14B, 508 one-to-one paired original questions) gives GAP +7.48 / +7.25 and D minus A -7.92 / -7.24 under question / database splits.}
\label{tab:robustness}
\scriptsize
\setlength{\tabcolsep}{3pt}
\begin{tabular}{p{0.36\linewidth}cccc}
\toprule
Check & Kwai-32B & Kwai-14B & Omni-32B & XiYan-32B \\
\midrule
instance cross-fit GAP / $\DmA$ (24 runs) & +9.81 / -9.88 & +7.84 / -7.96 & +8.91 / -8.51 & +3.11 / -2.90 \\
instance cross-fit, held-out D risk (range) & 0.094--0.101 & 0.097--0.099 & 0.096--0.101 & 0.100--0.101 \\
candidates-only partition GAP / $\DmA$ & +10.23 / -10.38 & +8.08 / -8.30 & +9.51 / -9.59 & +3.11 / -2.90 \\
gold-free construction GAP / $\DmA$ & +10.23 / -10.38 & +8.08 / -8.29 & +9.51 / -9.59 & +3.11 / -2.90 \\
exhaustive database splits, mean GAP / $\DmA$ & +10.35 / -10.66 & +8.10 / -8.18 & +9.61 / -10.19 & +2.81 / -1.39 \\
exhaustive splits, share with GAP $>0$ (range over seeds) & 100.0--100.0\% & 100.0--100.0\% & 100.0--100.0\% & 88.5--92.8\% \\
exhaustive splits, share with $\DmA$ $<0$ (range over seeds) & 99.9--100.0\% & 99.8--100.0\% & 100.0--100.0\% & 70.0--75.1\% \\
leave-one-out GAP, min / max & +8.96 / +10.80 & +6.97 / +8.95 & +8.34 / +11.98 & +1.17 / +3.86 \\
leave-one-out GAP, schema whose removal gives the min & concert\_singer & concert\_singer & concert\_singer & car\_1 \\
leave-one-out $\DmA$, weakest (schema removed) & -8.98 (concert\_singer) & -7.11 (concert\_singer) & -8.34 (concert\_singer) & -0.76 (car\_1) \\
\bottomrule
\end{tabular}
\end{table}

\begin{table}[htbp]
\centering
\caption{Where the D minus A contrast comes from, at $\alpha=0.10$, three-seed means over 200 splits. Both cells are judged by the same oracle, so a held-out question both cells answer with the same SQL contributes exactly zero and the three remaining events sum to the contrast; the identity is asserted on every split. The last column is the share of held-out questions on which both cells answer and the SQL is unchanged, which is not the 97.1 percent of Section~\ref{sec:populations}: that share counts the representative of every question, answered or not. Post hoc. File-order recomputation (\cref{app:tie}).}
\label{tab:dma_decomposition}
\footnotesize
\setlength{\tabcolsep}{5pt}
\begin{tabular}{llccccc}
\toprule
 & & & \multicolumn{3}{c}{contribution (pts)} & both answer, \\
\cmidrule(lr){4-6}
Checkpoint & Split & $\DmA$ (pts) & changed SQL & only D & only A & same SQL (\%) \\
\midrule
Kwai-32B & question & -10.380 & -0.001 & +0.000 & -10.379 & 74.1 \\
 & database & -10.695 & -0.001 & +0.023 & -10.717 & 74.9 \\
\addlinespace
Kwai-14B & question & -8.297 & +0.000 & +0.001 & -8.297 & 76.7 \\
 & database & -8.021 & +0.000 & +0.038 & -8.059 & 76.1 \\
\addlinespace
Omni-32B & question & -9.575 & -0.664 & +0.033 & -8.944 & 73.8 \\
 & database & -10.213 & -0.854 & +0.056 & -9.415 & 75.6 \\
\addlinespace
XiYan-32B & question & -2.902 & +0.005 & +0.031 & -2.938 & 91.8 \\
 & database & -1.246 & +0.026 & +0.338 & -1.610 & 85.8 \\
\bottomrule
\end{tabular}
\end{table}

\FloatBarrier
\section{Reproducibility}
\label{app:repro}

\paragraph{Code and data availability.} The sampling, evaluation and analysis code of this study is released at \dataurl.

The archive holds the scripts that draw the candidate pools, execute and compare queries under either oracle, build the two partitions, calibrate and evaluate the certificate, and carry out every analysis reported here: the panel entry rule, the oracle intervention and its aggregation, the alpha curves, the exhaustive database splits, the cross-fit, the gold-free construction, the six-score comparison, both censuses, both audit ladders, the decomposition of the $\DmA$ contrast, the reconstruction of the frozen tie convention, and the preregistered analyses of both expert audits together with their bounding and matched-population recomputations. The study's data is not in that archive: the per-question sufficient statistics of the twelve main and four control pools, the result files computed from them, the likelihood cache the two likelihood scores read, the two census label sets with their adjudications, and the two expert audit packages with their keys, sheets and preregistrations are held with the study and are not public. Every script names the inputs it reads and every quantity reported here was produced by running these scripts on exactly those inputs. The scripts that typeset the tables and draw the figures are released with them, and running those on the result files alone returns all fourteen generated tables byte for byte and produces all nine generated figures, the registration-history table of \cref{app:history} being written by hand and the schematic of \cref{fig:pipeline} drawn in the LaTeX source. The candidate pools themselves, about half a gigabyte of sampled SQL, are regenerable from the recorded prompts, sampling settings and seeds. The prompts and call traces of the AI labelling passes were not retained. The blind-audit item sheet is byte-reproducible from its key, whose content hash was recorded before any human label existed.

\paragraph{Environment.} Candidates were sampled with vLLM 0.22.1 in bfloat16 at a context of 8,192 tokens on NVIDIA H100 80GB cards, with tensor parallelism one for Kwai-14B and two for the three 32B checkpoints. Everything after sampling runs on CPU: a query is executed through the Python 3.12 \texttt{sqlite3} module against SQLite 3.45, and two results are compared by the official test-suite comparator of \citet{zhong2020testsuite} under the row-order convention of \cref{sec:system}. The released analysis scripts need only NumPy, and the scripts that typeset the tables and draw the figures also need Matplotlib.

\FloatBarrier
\section{The preregistered blinded human audit}
\label{app:humanaudit}

\begin{table}[htbp]
\centering
\caption{Result of the preregistered blinded human audit of the reference queries. Experts saw the question, the schema and the reference query only, executed the query, and returned one of four verdicts per item. $D$ is the share judged a defective reference query, with an exact binomial interval. The flagged stratum is a complete enumeration of the 73 flagged questions, so its interval describes uncertainty about a wider population of questions the flag could select rather than sampling error inside this set, while the two control intervals are sampling intervals for their own finite strata. The decision rule and its thresholds were committed before any human label existed.}
\label{tab:blind}
\small
\begin{tabular}{lrrrrr}
\toprule
Stratum & Items & Defective & $D$ (\%) & 95\% CI & Underspecified \\
\midrule
flagged & 73 & 40 & 54.8 & [42.7, 66.5] & 15 \\
examined, not flagged & 40 & 8 & 20.0 & [9.1, 35.6] & 9 \\
never examined & 40 & 3 & 7.5 & [1.6, 20.4] & 7 \\
\midrule
Preregistered verdict & \multicolumn{5}{p{0.70\linewidth}}{supported, margin 34.8 points} \\
Excluding borderline & \multicolumn{5}{p{0.70\linewidth}}{supported, margin 44.0 points on the first expert's marks, 60.2 if either expert's mark excludes} \\
Never-examined stratum & \multicolumn{5}{p{0.70\linewidth}}{7.5\% of its 40 sampled questions, 95\% CI 1.6 to 20.4} \\
Human versus AI flag & \multicolumn{5}{p{0.70\linewidth}}{agreement 71.2\%, $\kappa = 0.416$, over all 153 items} \\
Between experts & \multicolumn{5}{p{0.70\linewidth}}{binary agreement 94.1\%, $\kappa = 0.863$; thirteen four-way verdict disagreements adjudicated, nine of them binary} \\
\bottomrule
\end{tabular}
\end{table}

The human audit tests the suspected reference-defect label. All 73 flagged questions were included, together with two stratified random controls of 40 questions each that measure different things: the 130 questions whose rejected answers the census examined without assigning a reference-defect label test whether the flag discriminates, and the 305 questions none of whose answers the suite ever rejected measure the defect rate where these four checkpoints exposed nothing, which is the stratum a model-exposed count is silent about. Experts saw the question, the schema and the reference query only, in a shuffled order that carried no label vocabulary, stratum or question identifier, and were required to execute the reference query rather than read it, because the defect families of \cref{sec:c7} are execution behaviour. There was deliberately no contrast pass showing model answers: the never-examined stratum has no suite-rejected answer by construction, so showing one where it exists would reveal the stratum. Two experts labelled independently and only their disagreements were adjudicated, by a third pass that saw both verdicts and both rationales and was free to return a verdict neither expert had given. The decision rule was fixed and committed before any human label existed: the flag is supported if the flagged stratum's defect rate exceeds the examined stratum's by at least 25 points and reaches 40 percent, refuted if the difference is under 10 points, and partially supported otherwise; on refutation the suspected-defect interpretation of this paper would have been withdrawn and only the oracle-relative sensitivity results retained. Secondary analyses, also preregistered, are the background rate with an exact binomial interval, agreement between human and AI labels, confirmation rates by the strength of the AI label, and the sensitivity of every primary quantity to including or excluding borderline verdicts.

\Cref{tab:blind} reports the outcome. The two experts agreed on 94.1 percent of the 153 items on the binary defective-or-not reading and on 91.5 percent across all four verdicts, with Cohen's $\kappa$ of 0.863 on both readings. The third pass resolved their thirteen four-way disagreements, nine of which are also binary, ruling ten as the first expert had read them, two as the second had, and one as neither had. Under the adjudicated labels the flagged stratum was judged defective at 54.8 percent against 20.0 percent for the examined control, a margin of 34.8 points where the rule asks for 25, so the rule returns supported; dropping the verdicts marked borderline raises it further. That exclusion is convention-dependent, because twelve items carry the same verdict from both experts and different borderline marks, and the analysis reads the first expert's mark on them: it leaves 50 flagged and 25 examined items at a margin of 44.0 points, and excluding an item whenever either expert marked it leaves 42 and 22 at 60.2 points. Both clear the rule, and neither convention was preregistered.

In the never-examined stratum 3 of the 40 sampled questions carry a defective reference, 7.5 percent with an exact interval of 1.6 to 20.4. That stratum is defined by what the four checkpoints did not expose, so the figure is the defect rate among questions no answer of which the suite ever rejected, not a benchmark-wide prevalence: a defective reference is less likely to be caught where nothing was rejected. Weighting the three strata by their sizes gives the benchmark-wide point estimate, 17.5 percent over all 508 questions, which is post hoc and carries no preregistered interval. It is a stratified estimate in which only the two controls contribute sampling error, the flagged stratum being enumerated in full, and carries a design-based standard error of 2.7 points, an interval of 12.2 to 22.8 percent, three quarters of that variance coming from the never-examined stratum.

The human verdict and the AI flag agree on 71.2 percent of the 153 items, $\kappa = 0.416$. That figure counts the never-examined stratum as an AI negative, though the AI never looked at it; restricted to the 113 questions the AI did examine, agreement is 63.7 percent with $\kappa = 0.305$. Either way it is below what the two experts reach with each other, so the flag discriminates without being the same judgement. Confirmation tracks the strength of the AI label: 91.2 percent of the flagged questions whose label both AI passes assigned without marking it borderline were confirmed, against 27.6 percent of those the two passes merely agreed on and 10.0 percent of the ten that only the AI adjudication produced. Of the 51 references judged defective, the experts assigned a case-sensitive literal to 9, a wrong or negated filter to 7, numeric comparison or ordering on a text column to 7, a wrong grouping to 5, and two each to a wrong extremum, a missing join condition, an incomplete projection and a set operation keyed on a name; the remaining 15 fell outside the eight named families. The four families the census reports most often are among the four the experts used most often. Experts returned the underspecified-question verdict on 31 of the 153 items, 15 of them flagged, which is the preregistered secondary check on that reading of the questions; it is a judgement about a question alone, not about the point on which two queries differ, so it does not transfer to the census label of the same name.

Both experts are database engineers at tertiary hospitals with more than five years of SQL experience. They were recruited inside the laboratory and had not encountered this study before the package reached them: not the benchmark, not a model output, not a label of the AI audit, not the preregistration. They share a laboratory with the authors. The third pass over the thirteen disagreements was performed by the project lead. The preregistered category does not rest on it: run either expert's own sheet through the rule alone and it returns supported, at 36.2 points on the first and 27.7 on the second, with flagged rates of 56.2 and 45.2 percent against a threshold of 40. What the third pass decides is where inside that range the reported numbers fall. Its reach is bounded on both sides: no resolution of those thirteen items reaches refutation, and because the protocol lets the adjudicator return a verdict neither reviewer gave, the attainable margin runs from 20.2 to 38.7 points, so a reader who discounts the pass entirely still has partial support as the floor. Resolving the one item as neither reviewer had it lowered the margin rather than raising it.

Four limits bound what this establishes. The audit tested one label and tested it at the question level: the census attributes a defect to a particular reference query paired with a particular returned answer, and the experts saw no returned answer, so the 213 case-level attributions and the 887 answer occurrences they cover are supported only through the question-level flag. The semantic-error, underspecified-question, synthetic-instance and comparator-artefact labels of \cref{sec:census} remain AI-assigned. The experts were blind to the strata and to every model output but not to the hypothesis, and the protocol illustrated the defect families with families the AI audit itself had produced, each of which recurs among the flagged items. The protocol names no item and states that the families need not occur in the batch. And the protocol asks what a reference query returns on the supplied database, while six adjudications call a reference defective whose result on that database happens to be right and whose logic would fail on other data; scoring those six the other way gives 47.9 percent against 17.5 percent, a margin of 30.4 points, which the rule still supports.

\FloatBarrier
\section{The preregistered independent-yardstick audit}
\label{app:yardstick}

\begin{table}[t]
\centering
\caption{The preregistered independent-yardstick audit. Two SQL experts, blind to the cell, the score and every oracle label, judged whether each returned query answers its question, on the 1,028 distinct answers of two checkpoints; they were required to execute rather than read. AUROC is for predicting a correct answer, for the score built on the original-database partition and for the same score built on the suite partition, against the same expert labels on the same questions. The contrast is paired and tested by DeLong. The experts agree on 971 of the items; the last row of each block is the widest range a hill-climbing search over assignments of the remaining 57 attained; significance does not change anywhere inside it. The decision rule was fixed before any label existed and returns MIXED.}
\label{tab:alignment_audit}
\footnotesize
\setlength{\tabcolsep}{5pt}
\begin{tabular}{llccccc}
\toprule
Checkpoint & Labels & correct/wrong & AUROC on A & AUROC on D & $\Delta$ (pts) & $p$ \\
\midrule
XiYan-32B & both experts agree & 370/93 & 0.560 & 0.575 & +1.53 & 0.275 \\
 & first expert alone & 378/105 & 0.561 & 0.583 & +2.21 & 0.121 \\
 & second expert alone & 385/96 & 0.555 & 0.567 & +1.19 & 0.379 \\
 & searched assignments & & & & +0.47 to +2.72 & 0.076 to 0.688 \\
\addlinespace
Kwai-32B & both experts agree & 364/91 & 0.627 & 0.697 & +6.96 & 0.004 \\
 & first expert alone & 377/99 & 0.632 & 0.702 & +7.08 & 0.002 \\
 & second expert alone & 373/95 & 0.628 & 0.689 & +6.09 & 0.009 \\
 & searched assignments & & & & +4.18 to +8.83 & 0.000 to 0.045 \\
\bottomrule
\end{tabular}
\end{table}

\begin{table}[t]
\centering
\caption{The certificate re-scored against the expert labels of \cref{app:yardstick}, at $\alpha=0.10$ under question splits, generation seed 101. Unlike every earlier relabelling, these labels cover both sides of the suite oracle, so the sign is read off rather than bracketed. GAP is the risk the current-practice cell carries under the experts minus the risk it reports on its own labels; D minus A is the expert-judged risk of the fully multi-instance cell minus that of the current-practice cell, each cell answering at the threshold its own calibration selected. Rows differ in what is done with the questions the experts called underspecified or could not judge. Post hoc: the preregistration for this audit fixed the alignment endpoint and declined to preregister a claim about the sign.}
\label{tab:human_gap}
\footnotesize
\setlength{\tabcolsep}{5pt}
\begin{tabular}{llcc}
\toprule
Checkpoint & Unjudged verdicts & GAP (pts) & $\DmA$ (pts) \\
\midrule
XiYan-32B & the suite label & +10.10 & -2.42 \\
 & the shipped-database label & +9.25 & -1.86 \\
 & dropped & +10.14 & -1.42 \\
 & either expert alone & +9.41 to +11.93 & -1.49 to -2.52 \\
 & \multicolumn{3}{l}{\footnotesize experts call wrong 17.7\% of the 407 answers the suite accepts, 36.8\% of the 57 it rejects} \\
\addlinespace
Kwai-32B & the suite label & +7.37 & -4.57 \\
 & the shipped-database label & +6.19 & -3.79 \\
 & dropped & +6.79 & -3.58 \\
 & either expert alone & +6.42 to +7.94 & -3.75 to -4.41 \\
 & \multicolumn{3}{l}{\footnotesize experts call wrong 15.1\% of the 352 answers the suite accepts, 36.3\% of the 113 it rejects} \\
\bottomrule
\end{tabular}
\end{table}

\Cref{sec:alignment} leaves open whether the suite partition builds a better score or only looks better under the labels it produced, and neither oracle can settle that, so the audit brings in a third reading. Two SQL experts saw a question, its schema and one returned query, with no reference query, no score, no cell, no oracle label and no stratum, in an order shuffled under a fixed seed, and were required to execute the query rather than read it. The four verdicts are that the query answers the question, that it does not, that the question is underspecified so that both readings are faithful to it, and that the item cannot be judged; the last two are excluded from the endpoint and their share is reported. They cover 4.5 and 5.4 percent of the two sheets, and counting them as correct instead moves the contrast to 0.96 points on XiYan-32B and 7.08 on Kwai-32B, changing neither significance.

The endpoint is confined to the questions on which both cells return the same SQL, 507 of 508 on XiYan-32B and 496 of 508 on Kwai-32B, because there one verdict serves both cells and the cells differ only in the score. The 26 items covering the 13 questions where the returned SQL differs were judged as well and are reported separately: the experts agreed on 23 of them, 16 as answering the question and 5 as not. The sheet holds distinct answers rather than checkpoint-and-cell uses, and it deduplicates across checkpoints as well: on one question both checkpoints return the same SQL, so that answer was judged once and serves four uses, which is why the same-SQL items number 1,002 rather than 1,003 and the sheet totals 1,028. The endpoint then drops the items the experts disagree on and those whose verdict is not usable, so it rests on 463 of those questions on XiYan-32B and 455 on Kwai-32B (\cref{tab:alignment_audit}), and every suite-label contrast quoted against it is recomputed on those same questions: over all same-SQL questions the suite labels put the suite-partition score 9.2 points ahead on XiYan-32B and 8.9 on Kwai-32B, and on the questions the expert estimate retains those contrasts are 8.3 and 9.2, so the comparison in \cref{sec:yardstick} changes the label without changing the population.

Two things were fixed before any label existed. The decision rule declares the outcome substantive when the suite-partition score has the higher AUROC on both checkpoints at $p < 0.05$, circular when neither does, and mixed otherwise. Those three names are the labels the registration gave to outcome patterns, and not mechanisms this design identifies: a non-detection bounds the advantage on that checkpoint rather than establishing that the score carries none, so what the rule returns is read here as persistence or non-detection under an independent yardstick. And the two checkpoints were picked as opposite cases under the oracles' own labels, one symmetric and one not. The advantage persisted on the symmetric checkpoint, Kwai-32B, and was not detected on the other, XiYan-32B.

The audit differs from that of \cref{app:humanaudit} in two respects: neither expert took part in that earlier round, and the protocol illustrates no defect family. The design calculation, run before the item sheet was frozen, put the paired standard error at 0.024 and the separable contrast at about 4.5 points; the realised standard errors are 0.014 and 0.024, so the audit resolves a contrast of that size, and on XiYan-32B the interval of \cref{sec:yardstick}, which runs to 4.3 points, excludes the 8.3 points the suite labels report. The 57 disagreements are bracketed by search rather than adjudicated, and the verdict does not change across the range the search attained.

\paragraph{Two preregistered secondary analyses of the mechanism.} The first substitutes the expert labels for the suite's in the reversal test of \cref{sec:alignment}, where the weak-minus-suite change in AUROC is positive for the score built on the shipped-database partition and negative for the score built on the suite partition, in 16 of 16 combinations. Replacing the suite labels by the expert labels removes the reversal on both audited checkpoints: the weak-minus-expert change is positive for both cells, 9.7 and 12.9 points on XiYan-32B and 22.0 and 12.9 on Kwai-32B, because the expert labels are harder for both scores and come from neither score's own oracle. The reversal is therefore a property of the relationship between the two oracles, not evidence that either partition builds the better score, which is the separate question \cref{tab:alignment_audit} answers.

The second bounds the instance-level component of the alignment without any new label. The archived cross-fit pools build the suite partition and its labels from one half of each schema's suite instances and judge correctness on the other half, so a score can be read under the instances that built it and under instances it never saw. The suite-partition score loses more than the shipped-database score in six of the eight pools, the direction that alignment predicts, but the losses are 0.2 to 1.3 points of AUROC against an alignment effect near 9 points, and on XiYan-32B they are exactly zero because the two instance halves disagree on the label of one or two classes in about 1,255. Instance-level circularity, as this cross-fit measures it, is present and an order of magnitude smaller than the alignment. What remains is carried by whatever the two halves share; the reference queries are the candidate explanation this design cannot separate from the benchmark's other shared assumptions, and an expert audit reaches that shared component where cross-fitting cannot.

\end{document}